\documentclass[11pt]{article}

\usepackage[final]{acl}

\usepackage{times}
\usepackage{latexsym}
\usepackage{amsmath}
\usepackage[T1]{fontenc}

\usepackage[utf8]{inputenc}
\usepackage{graphicx}

\usepackage{microtype}
\usepackage{amsmath}
\usepackage{inconsolata}

\usepackage{amssymb} 
\usepackage{booktabs}
\usepackage{multirow}
\usepackage{xcolor}
\usepackage{hyperref}
\usepackage{url}

\title{DPN-LE: Dual Personality Neuron Localization and Editing for Large Language Models}

\author{
 \textbf{Lifan Zheng\textsuperscript{1}},
 \textbf{Xue Yang\textsuperscript{2}},
 \textbf{Jiawei Chen\textsuperscript{3,4}},
 \textbf{Chenyan Wu\textsuperscript{5}},
\\
 \textbf{Jingyuan Zhang\textsuperscript{6}},
 \textbf{Fanheng Kong\textsuperscript{7}},
 \textbf{Xinyi Zeng\textsuperscript{8}},
 \textbf{Xiang Chen \textsuperscript{9}},
\\
 \textbf{Yu Tian\textsuperscript{8\thanks{Corresponding Author.}}}
\\
\\
 \textsuperscript{1}Southeast University,
 \textsuperscript{2}Shanghai Jiao Tong University
 \textsuperscript{3}East China Normal University
 \textsuperscript{4}Zhongguancun Academy
\\
 \textsuperscript{5}Zhejiang University of Technology
 \textsuperscript{6}Kuaishou Technology
 \textsuperscript{7}Northeastern University
\\
 \textsuperscript{8}Tsinghua University
 \textsuperscript{9}Nanjing University of Aeronautics and Astronautics
\\
 \small{
   \textbf{Correspondence:} 
   \href{mailto:z1ivan@seu.edu.cn}{z1ivan@seu.edu.cn},
   \href{mailto:tianyu181@mails.ucas.ac.cn}{tianyu181@mails.ucas.ac.cn}
 }
}

\begin{document}
\maketitle

\begin{abstract}
With the widespread adoption of large language models (LLMs), understanding their personality representation mechanisms has become critical. As a novel paradigm in Personality Editing, most existing methods employ neuron-editing to locate and modify LLM neurons, requiring changes to numerous neurons and leading to significant performance degradation. This raises a fundamental question: Are all modified neurons directly related to personality representation? In this work, we investigate and quantify this specificity through assessments of general capability impact and representation-level patterns. We find that: 1) Current methods can change personalities but reduce overall performance. 2) Neurons are multifunctional, connecting personality traits and general knowledge. 3) Opposing personality traits demonstrate distinctly mutually exclusive representation patterns. Motivated by these findings, we propose DPN-LE (Dual Personality Neuron Localization and Editing), which identifies personality-specific neurons by contrasting MLP activations between high-trait and low-trait samples. DPN-LE constructs layer-wise steering vectors and applies dual-criterion filtering based on Cohen's $d$ effect size and activation magnitude to isolate mutually exclusive neuron subsets. Sparse linear intervention on these neurons enables precise personality control at inference time. Using only 1,000 contrastive sample pairs per trait, DPN-LE intervenes on $\sim$0.5\% of neurons while achieving competitive personality control and substantially better capability preservation across reasoning tasks. Experiments on LLaMA-3-8B-Instruct and Qwen2.5-7B-Instruct demonstrate the effectiveness and generalizability of our approach \footnote[1]{Code: \url{https://github.com/Z1ivan/DPN-LE}}.
\end{abstract}

\section{Introduction}
With the rapid development of large language models (LLMs), understanding their personality representation mechanisms has become a critical research focus, providing technical support for applications such as social surveys, role-playing, and personality analysis \cite{park2023generative,shao2023character,wang2024rolellm,cao2024large,chen2025red,wang2025deeppersona}. These scenarios demand that models fulfill dual objectives: possessing robust reasoning for logical consistency and exhibiting nuanced personality traits for natural interaction. Therefore, understanding and editing personality traits in LLMs is essential for building responsive and adaptable LLMs.

\begin{figure}[t]
    \centering
    \includegraphics[width=0.9\columnwidth]{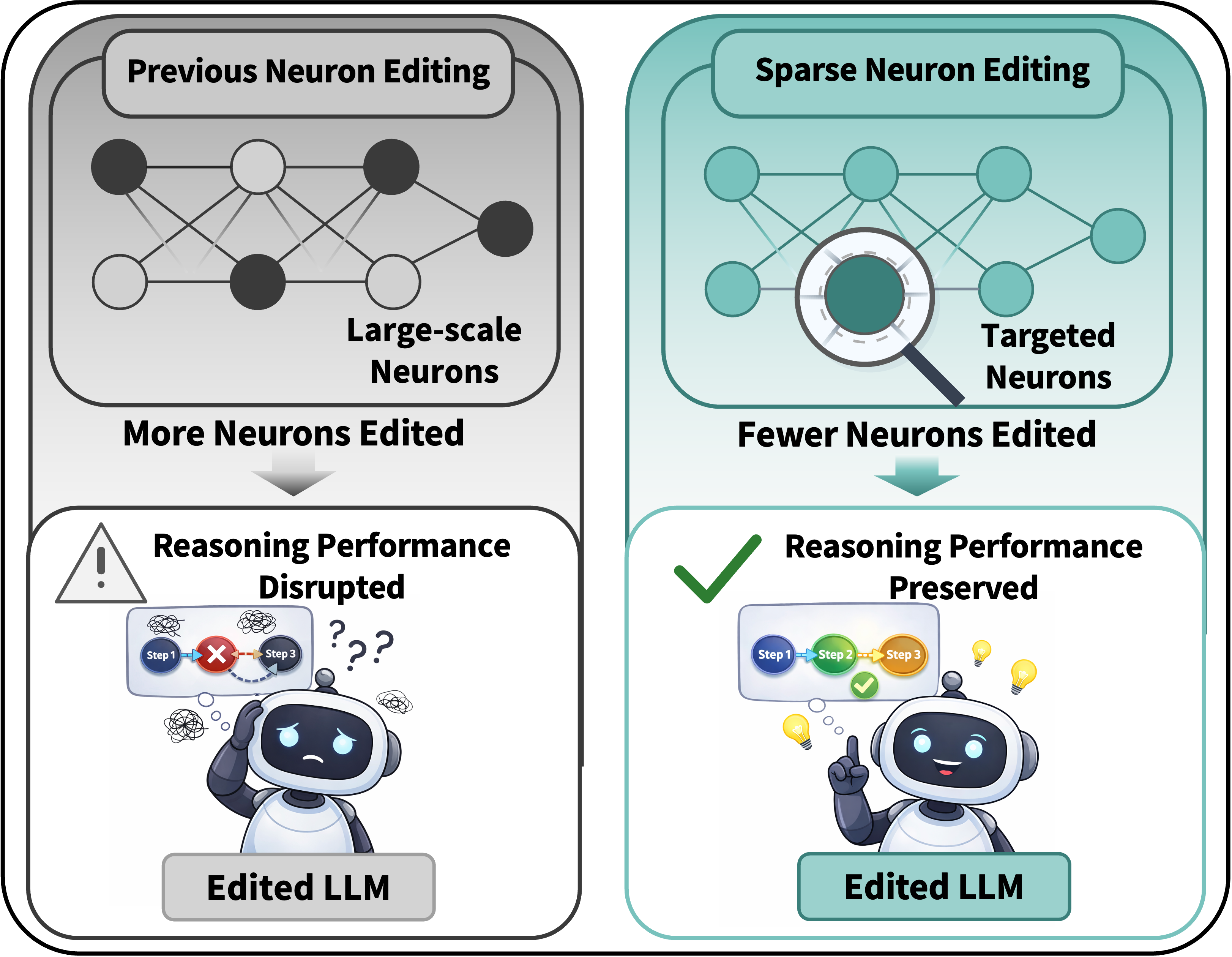}
    \caption{Comparison between previous large-scale neuron editing and our sparse personality-specific editing.}
    \label{fig:intro}
\end{figure}

\begin{table*}[ht]
\centering
\small
\begin{tabular}{l|cc|cc|cc|cc|cc}
\toprule
 & \multicolumn{2}{c|}{\textbf{GSM8K Acc}} & \multicolumn{4}{c|}{\textbf{HotpotQA}} & \multicolumn{4}{c}{\textbf{TriviaQA}} \\
 & \multicolumn{2}{c|}{(Baseline: 75.36)} & \multicolumn{2}{c|}{EM (B: 13.0)} & \multicolumn{2}{c|}{F1 (B: 25.24)} & \multicolumn{2}{c|}{EM (B: 66.4)} & \multicolumn{2}{c}{F1 (B: 60.60)} \\
\cmidrule{2-11}
\textbf{Trait} & \textbf{+} & \textbf{-} & \textbf{+} & \textbf{-} & \textbf{+} & \textbf{-} & \textbf{+} & \textbf{-} & \textbf{+} & \textbf{-} \\
\midrule
Openness & -17.59 & -66.03 & +1.0 & -1.9 & -0.78 & -3.65 & -6.6 & -11.5 & -4.83 & -6.85 \\
Conscientiousness & -5.16 & -29.34 & +3.1 & +1.7 & +1.45 & -0.08 & -1.8 & -4.3 & -1.35 & -2.82 \\
Extraversion & -14.56 & -60.27 & +3.9 & -2.3 & +0.73 & -4.70 & -8.0 & -7.0 & -5.50 & -5.10 \\
Agreeableness & -15.09 & -33.21 & -4.0 & -0.6 & -3.82 & -1.84 & -5.4 & -5.7 & -4.15 & -2.91 \\
Neuroticism & -27.60 & -15.09 & -1.6 & -4.2 & -2.78 & -3.78 & -3.8 & -3.8 & -2.25 & -4.00 \\
\midrule
\textbf{Average} & -16.00 & -40.79 & +0.48 & -1.46 & -1.04 & -2.81 & -5.12 & -6.46 & -3.61 & -4.34 \\
\bottomrule
\end{tabular}
\caption{General capability degradation with NPTI ($\gamma=1.4$) on LLaMA-3-8B-Instruct. + and - denote personality high-trait and low-trait directions, respectively.}
\label{tab:npti-general}
\end{table*}

\begin{figure*}[t]
    \centering
    \includegraphics[width=\textwidth]{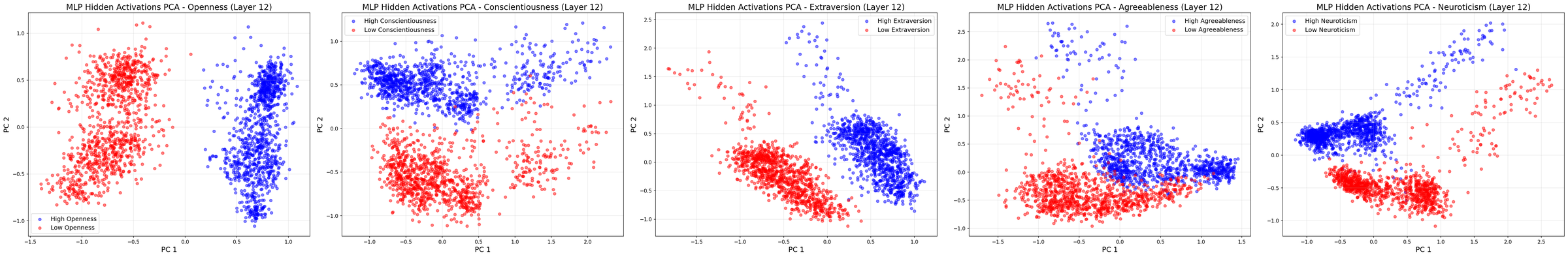}
    \caption{PCA visualization of MLP activations at Layer 12 for all Big Five traits on LLaMA-3-8B-Instruct. Red points represent high-trait samples, blue points represent low-trait samples. Clear separation between opposing traits emerges at this layer.}
    \label{fig:pca-layer12}
\end{figure*}

Current methods for editing personality traits in LLMs can be divided into two categories. \textit{Prompt-based methods} induce personality by modifying system prompts. While these methods can quickly induce personality traits, they heavily rely on prompt design and exhibit limitations in stability and persistence \cite{huang2023humanity,serapio2023personality}. \textit{Neuron-editing methods} achieve precise intervention by locating and editing neurons that influence personality representations \cite{meng2022locating,meng2022mass,deng2024neuron}. However, these methods suffer from significant performance degradation due to numerous neurons modified. This dilemma raises a critical question: \textbf{Are all modified neurons directly related to personality representation?}

To address this question, we systematically evaluate the impact of existing neuron-editing methods on model performance and their redundancy. First, we analyze changes in the general capabilities of LLMs, including mathematical reasoning and question answering. Then, we employ Principal Component Analysis (PCA) to characterize the activation patterns at the internal representation level. Our findings reveal that: (1) current methods effectively alter personalities but substantially degrade general performance; (2) neurons exhibit multifunctionality, being associated with both personality traits and general knowledge; and (3) opposing personality traits manifest as markedly mutually exclusive patterns in the representation space.

\begin{figure*}[t]
    \centering
    \includegraphics[width=\textwidth]{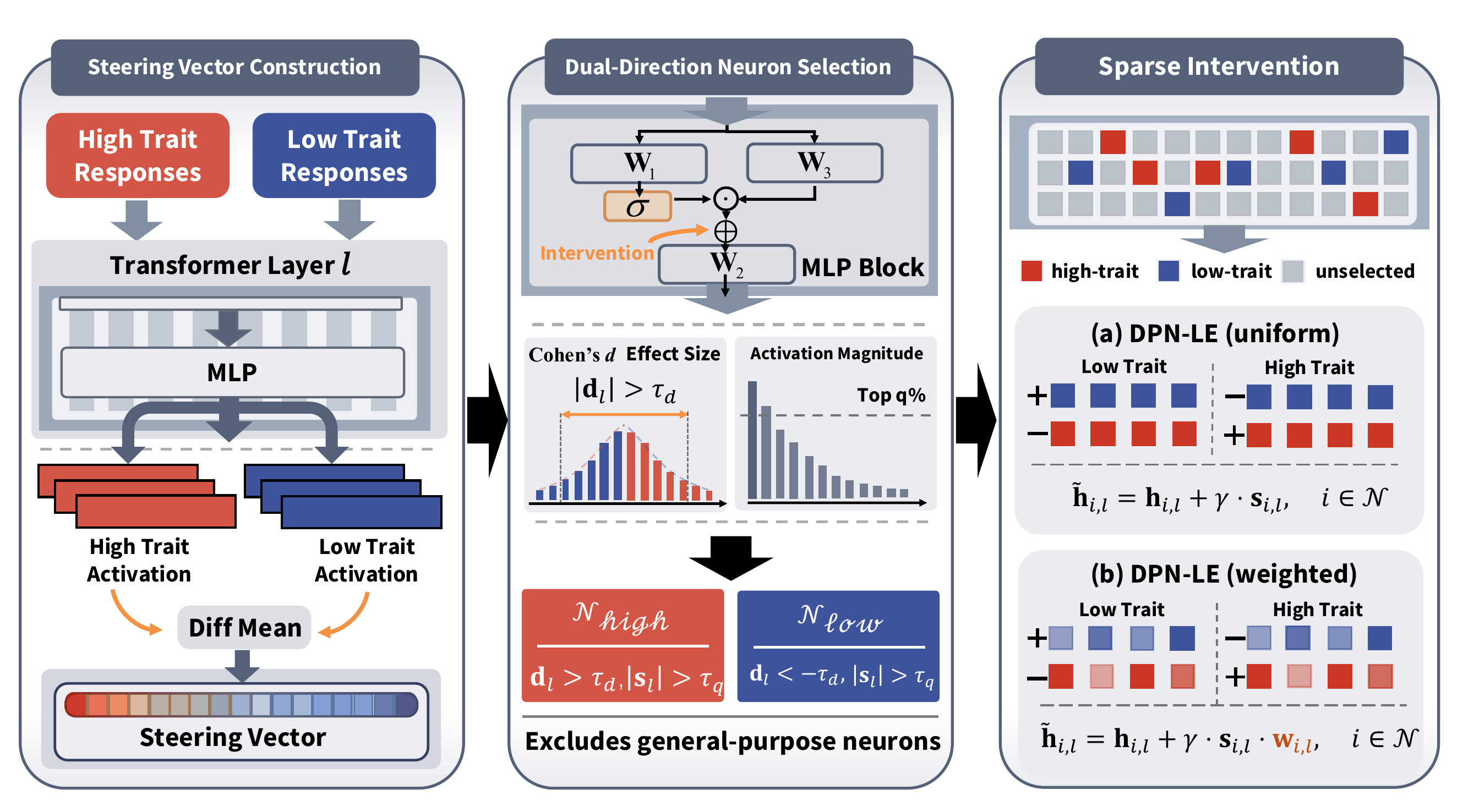}
    \caption{Overview of DPN-LE. (1) We construct steering vectors by computing the mean activation difference between high-trait and low-trait samples. (2) We apply dual-criterion filtering (Cohen's $d$ threshold and quantile threshold) to select trait-exclusive neurons. (3) During inference, we apply sparse interventions only to the selected neurons for precise personality control.}
    \label{fig:method}
\end{figure*}

Motivated by these findings, we propose \textbf{DPN-LE} (\textbf{D}ual-\textbf{P}ersonality-\textbf{N}euron \textbf{L}ocalization and \textbf{E}diting), which identifies personality-related neurons by contrasting activation patterns between opposing personality traits. As shown in Figure \ref{fig:method}, DPN-LE constructs layer-wise steering vectors from MLP activations and applies dual-direction filtering based on effect size to identify trait-exclusive neuron subsets. During inference, sparse linear interventions on hidden representations enable precise personality control without modifying model weights. Extensive experiments demonstrate that DPN-LE achieves competitive personality control by intervening on only 0.5\% of neurons, while substantially better preserving general reasoning capabilities. Our main contributions are as follows:

\begin{itemize}
    \item We systematically evaluate neuron-editing methods and reveal substantial redundancy in modified neurons, with many neurons being unrelated to personality representation.
    \item We propose DPN-LE, which leverages the mutual exclusivity between opposing personality traits to precisely localize personality-related neurons, reducing modified parameters by over 90\% compared to existing methods.
    \item We design two intervention strategies (DPN-LE and DPN-LE$_w$) that achieve stable personality control by intervening on only 0.5\% of neurons while maintaining minimal impact on general reasoning capabilities.
\end{itemize}

\section{Related Work}
\noindent\textbf{Personality in LLMs}
Research on personality in LLMs spans assessment, induction, and consistency. For assessment, researchers have adapted psychological instruments such as the Big Five model \cite{mccrae1992introduction,digman1990personality,goldberg1999broad} and MBTI \cite{pan2023llms} to evaluate LLM personality traits \cite{jiang2024personallm}. For induction, prompt-based methods like $P^2$ \cite{jiang2023evaluating} design personality-descriptive prompts, while fine-tuning approaches train on personality-annotated dialogues \cite{li2023chatharuhi}. However, studies reveal that LLMs often exhibit inconsistent personality across different contexts \cite{dorner2023personality}, motivating the need for more robust control mechanisms at the representation level.

\noindent\textbf{Neuron Localization and Editing}
Neuron localization methods identify neurons associated with specific behaviors. Knowledge Neurons trace factual knowledge to specific MLP neurons through gradient-based attribution \cite{dai2022knowledge}. ROME and MEMIT \cite{meng2022locating,meng2022mass} further develop causal tracing techniques to locate and edit factual associations in MLP layers. For personality, NPTI \cite{deng2024neuron} identifies neurons that activate differently for high versus low trait expressions by computing activation probability differences. However, NPTI typically modifies tens of thousands of neurons per trait, which may also affect neurons involved in general reasoning, rather than only trait-specific ones \cite{mu2020compositional,bau2020understanding}.

\noindent\textbf{Activation Steering}
Activation steering controls model behavior by adding steering vectors to internal representations during inference \cite{zou2023representation,turner2024activation}. CAA \cite{rimsky2024steering} constructs steering vectors by averaging activation differences between contrastive examples, then adds them to the residual stream at all token positions \cite{li2023inference}. PAS \cite{zhu2024personality} identifies effective attention heads and optimizes activation offsets for personality alignment. However, these methods lack neuron-level selection within target components, potentially affecting neurons unrelated to personality and causing unnecessary interference with general capabilities.

\begin{table*}[t]
\resizebox{\textwidth}{!}{
\centering
\small
\begin{tabular}{c|c|cc|cc|cc|cc|cc|cc}
\toprule
\multirow{2}{*}{Metric} & \multirow{2}{*}{Trait} & \multicolumn{2}{c|}{Simple Prompt} & \multicolumn{2}{c|}{$P^2$} & \multicolumn{2}{c|}{PAS} & \multicolumn{2}{c|}{DPN-LE} & \multicolumn{2}{c|}{DPN-LE$_w$} & \multicolumn{2}{c}{{\color[HTML]{656565} NPTI}} \\
\cmidrule{3-14}
& & mean$\uparrow$ & var$\downarrow$ & mean$\uparrow$ & var$\downarrow$ & mean$\uparrow$ & var$\downarrow$ & mean$\uparrow$ & var$\downarrow$ & mean$\uparrow$ & var$\downarrow$ & {\color[HTML]{656565} mean$\uparrow$} & {\color[HTML]{656565} var$\downarrow$} \\
\midrule
\multirow{6}{*}{\rotatebox{90}{Trait}}
& Agreeableness     & \textbf{9.73} & \textbf{0.32} & 9.59 & 0.47 & 6.48 & 1.01 & 9.47 & 0.57 & 9.56 & 0.48 & {\color[HTML]{656565} 9.64} & {\color[HTML]{656565} 0.49} \\
& Conscientiousness & 8.98 & 0.98 & 9.09 & 0.82 & 6.69 & 1.63 & 8.75 & 0.83 & 8.64 & 0.81 & {\color[HTML]{656565} \textbf{9.25}} & {\color[HTML]{656565} \textbf{0.66}} \\
& Extraversion      & 9.26 & 1.04 & 9.39 & 0.64 & 7.57 & 2.81 & 8.82 & 1.32 & 8.79 & 1.20 & {\color[HTML]{656565} \textbf{9.86}} & {\color[HTML]{656565} \textbf{0.14}} \\
& Neuroticism       & 8.04 & 1.18 & 9.53 & 0.55 & 6.98 & 1.58 & 9.93 & 0.07 & \textbf{9.95} & \textbf{0.05} & {\color[HTML]{656565} 9.92} & {\color[HTML]{656565} 0.07} \\
& Openness          & 7.42 & 1.86 & \textbf{9.16} & 0.72 & 6.93 & 1.52 & 8.58 & \textbf{0.74} & 8.58 & 0.80 & {\color[HTML]{656565} 8.50} & {\color[HTML]{656565} 1.08} \\
& Average           & 8.69 & 1.08 & 9.35 & 0.64 & 6.93 & 1.71 & 9.11 & 0.71 & 9.10 & 0.67 & {\color[HTML]{656565} \textbf{9.43}} & {\color[HTML]{656565} \textbf{0.49}} \\
\midrule
\multirow{6}{*}{\rotatebox{90}{Fluency}}
& Agreeableness     & 9.36 & 0.56 & 9.76 & \textbf{0.23} & \textbf{9.83} & 0.27 & 8.98 & 0.99 & 9.01 & 0.73 & {\color[HTML]{656565} 9.72} & {\color[HTML]{656565} \textbf{0.23}} \\
& Conscientiousness & 9.89 & 0.12 & 9.92 & 0.07 & 9.92 & 0.07 & 9.04 & 0.76 & 9.00 & 0.67 & {\color[HTML]{656565} \textbf{9.96}} & {\color[HTML]{656565} \textbf{0.04}} \\
& Extraversion      & 9.99 & 0.01 & \textbf{10.00} & \textbf{0.00} & 9.98 & 0.02 & 9.02 & 1.05 & 8.81 & 1.26 & {\color[HTML]{656565} 9.88} & {\color[HTML]{656565} 0.11} \\
& Neuroticism       & \textbf{10.00} & \textbf{0.00} & \textbf{10.00} & \textbf{0.00} & \textbf{10.00} & \textbf{0.00} & 9.80 & 0.32 & 9.71 & 0.43 & {\color[HTML]{656565} 9.91} & {\color[HTML]{656565} 0.09} \\
& Openness          & 9.91 & 0.08 & 9.83 & 0.16 & \textbf{9.97} & \textbf{0.03} & 8.56 & 1.00 & 8.61 & 0.97 & {\color[HTML]{656565} 9.83} & {\color[HTML]{656565} 0.18} \\
& Average           & 9.83 & 0.15 & \textbf{9.90} & 0.09 & \textbf{9.94} & \textbf{0.08} & 9.08 & 0.82 & 9.03 & 0.87 & {\color[HTML]{656565} 9.86} & {\color[HTML]{656565} 0.13} \\
\bottomrule
\end{tabular}%
}
\caption{Automatic evaluation results on LLaMA-3-8B-Instruct. The upper section shows personality trait scores, and the lower section shows fluency scores. Simple Prompt, $P^2$, DPN-LE, and DPN-LE$_w$ are reproduced by us. PAS and NPTI results are reported from \cite{deng2024neuron}.}
\label{tab:auto-eval-combined}
\end{table*}

\begin{table}[ht]
\centering
\small
\begin{tabular}{r|rr|rr|r}
\toprule
\multirow{2}{*}{\textbf{Trait}} & \multicolumn{2}{c|}{\textbf{NPTI}} & \multicolumn{2}{c|}{\textbf{DPN-LE}} & \multirow{2}{*}{\textbf{Reduction}} \\
\cmidrule{2-5}
& \textbf{+} & \textbf{-} & \textbf{+} & \textbf{-} & \\
\midrule
Ope. & 28,193 & 31,790 & 728 & 701 & 97.6\% \\
Cons. & 10,278 & 16,997 & 703 & 719 & 94.8\% \\
Ext. & 27,427 & 24,519 & 728 & 705 & 97.2\% \\
Agr. & 21,008 & 21,083 & 661 & 746 & 96.7\% \\
Neu. & 19,211 & 16,313 & 733 & 696 & 96.0\% \\
\midrule
\textbf{Average} & 21,223 & 22,140 & 711 & 713 & \textbf{96.7\%} \\
\bottomrule
\end{tabular}
\caption{Comparison of modified neuron counts between NPTI and DPN-LE. + and - denote high-trait and low-trait directions, respectively.}
\label{tab:neuron-comparison}
\end{table}

\section{Preliminary}

\subsection{Experimental Setup}
We conduct preliminary experiments on LLaMA-3-8B-Instruct \cite{grattafiori2024llama}. NPTI is applied to induce each of the Big Five personality traits in both high-trait (+) and low-trait (-) directions. To evaluate general capabilities, we use three benchmarks: GSM8K \cite{cobbe2021training} for mathematical reasoning (Accuracy), HotpotQA \cite{yang2018hotpotqa} for multi-hop question answering, and TriviaQA \cite{joshi2017triviaqa} for factual knowledge retrieval. For QA tasks, we report Exact Match (EM) and F1 score. 
NPTI identifies approximately 20,000 neurons per trait through the PersonalityBench dataset \cite{deng2024neuron}, then modifies their activation values in MLP layers during inference. Following the original settings, we use enhancement coefficient $\gamma=1.4$ with sigmoid-weighted modulation.

\subsection{Analysis}
\noindent\textbf{General Capability Degradation.}
As shown in Table~\ref{tab:npti-general}, we observe significant decline in general capabilities after personality editing. The baseline model achieves 75.36\% on GSM8K, but accuracy drops by 5.16\%--66.03\% after personality editing, with the low-trait direction causing more severe degradation. HotpotQA shows relatively stable performance with average EM changes of +0.48\% (high) and -1.46\% (low), and F1 drops of 1.04\% (high) and 2.81\% (low). TriviaQA exhibits moderate degradation with EM drops of 5.12\% (high) and 6.46\% (low), and F1 drops of 3.61\% (high) and 4.34\% (low).

These results suggest that current methods can effectively alter the personality of LLMs; however, \textbf{extensive neuron modifications lead to decreased general capabilities.} Notably, we find that the model's performance for the low-trait direction is significantly lower than for the high-trait direction. We believe this is due to the need for the model to inhibit its existing expressive patterns when suppressing personality traits, requiring more complex neural regulation. In contrast, enhancing personality traits amplifies current neural signals based on the existing state, as LLMs typically operate in a positive state without personality editing, resulting in less interference.

\noindent\textbf{Representation-Level Analysis.}
To further investigate the root reasons for the poor general capability of current methods, we conduct PCA analysis on MLP activations for both high-trait and low-trait samples across all layers. As shown in Figure~\ref{fig:pca-layer12}, opposing personality traits form clearly separable clusters starting from layer 12 for LLaMA-3-8B-Instruct. We find that: 1) \textbf{there exist trait-exclusive neurons that respond strongly to only one direction.} 2) \textbf{Neurons in the intersectional areas are multifunctional, relating to both personality traits and general knowledge.}

\noindent\textbf{Motivation.}
Based on the analysis of the pilot experiment, we observe that redundancy in current methods leads to the selection of numerous non-exclusive neurons, which interfere with general capabilities when modified. This observation motivates our dual-direction filtering approach, which selects only neurons with large effect sizes in one direction, effectively identifying a sparse subset of truly personality-specific neurons.

\begin{table*}[ht]
\centering
\small
\begin{tabular}{ll|cc|cc|cc|cc|cc}
\toprule
 & & \multicolumn{2}{c|}{\textbf{GSM8K Acc}} & \multicolumn{4}{c|}{\textbf{HotpotQA}} & \multicolumn{4}{c}{\textbf{TriviaQA}} \\
 & & \multicolumn{2}{c|}{(Baseline: 75.36)} & \multicolumn{2}{c|}{EM (B: 13.0)} & \multicolumn{2}{c|}{F1 (B: 25.24)} & \multicolumn{2}{c|}{EM (B: 66.4)} & \multicolumn{2}{c}{F1 (B: 60.60)} \\
\cmidrule{3-12}
\textbf{Method} & \textbf{Trait} & \textbf{+} & \textbf{-} & \textbf{+} & \textbf{-} & \textbf{+} & \textbf{-} & \textbf{+} & \textbf{-} & \textbf{+} & \textbf{-} \\
\midrule
\multirow{6}{*}{DPN-LE} & Openness & -15.62 & -5.00 & -0.4 & -2.3 & -1.80 & -4.34 & -2.8 & -9.1 & -2.21 & -5.74 \\
 & Conscientiousness & -5.38 & -1.67 & -1.5 & -2.1 & -2.26 & -2.72 & -1.9 & -8.8 & -2.17 & -5.54 \\
 & Extraversion & -5.99 & -13.19 & -4.0 & +1.2 & -5.89 & -0.63 & -11.8 & -8.7 & -7.85 & -5.81 \\
 & Agreeableness & -5.76 & +1.14 & -0.8 & -1.4 & -1.71 & -2.40 & -1.2 & -8.0 & -1.02 & -4.96 \\
 & Neuroticism & -18.80 & -6.97 & -1.3 & -2.3 & -3.32 & -3.63 & -13.5 & -6.3 & -8.50 & -4.26 \\
\cmidrule{2-12}
 & \textbf{Average} & -10.31 & -5.14 & -1.60 & -1.38 & -3.00 & -2.74 & -6.24 & -8.18 & -4.35 & -5.26 \\
\midrule
\multirow{6}{*}{DPN-LE$_w$} & Openness & -8.49 & -5.00 & +0.3 & -1.6 & -0.90 & -3.14 & -1.5 & -7.3 & -1.39 & -4.49 \\
 & Conscientiousness & -6.14 & -1.36 & -1.3 & -1.4 & -1.65 & -2.05 & -0.6 & -5.5 & -1.37 & -3.39 \\
 & Extraversion & -4.17 & -17.89 & -3.1 & +1.1 & -4.49 & -0.55 & -8.4 & -6.6 & -5.32 & -4.41 \\
 & Agreeableness & -5.23 & +0.38 & -0.7 & -1.2 & -1.29 & -2.34 & -0.1 & -5.3 & -0.49 & -3.44 \\
 & Neuroticism & -11.37 & -5.76 & -0.4 & -2.1 & -1.93 & -3.28 & -9.3 & -4.6 & -5.83 & -3.28 \\
\cmidrule{2-12}
 & \textbf{Average} & -7.08 & -5.93 & -1.04 & -1.04 & -2.05 & -2.27 & -3.98 & -5.86 & -2.88 & -3.80 \\
\bottomrule
\end{tabular}
\caption{General capability with DPN-LE and DPN-LE$_w$ ($\gamma=0.8$) on LLaMA-3-8B-Instruct. + and - denote personality high- and low-trait directions, respectively.}
\label{tab:dpn-general}
\end{table*}

\section{Methodology}

Based on the preliminary findings that trait-exclusive neurons exist and can be identified through activation contrasts, we propose DPN-LE (\textbf{D}ual-\textbf{P}ersonality-\textbf{N}euron \textbf{L}ocalization and \textbf{E}diting). Our approach consists of three stages: (1) constructing steering vectors from MLP activations, (2) selecting personality-exclusive neurons via dual-direction filtering, and (3) applying sparse interventions during inference. Figure~\ref{fig:method} illustrates the overall framework.

\subsection{Steering Vector Construction}
For a target personality trait (e.g., Neuroticism), we collect a dataset $\mathcal{D} = (x_i^+,x_i^-)_{i=1}^N$, where $x_i^+$ is the high trait and $x_i^-$ is the low trait. At each Transformer layer $l$, we extract the MLP hidden state (computed after the gated activation) at the last token position, denoted as  $\mathbf{H}_{i,l}^+$ and $\mathbf{H}_{i,l}^-$ for high and low trait. All hidden states $\mathbf{H}_{i,l}$ include $K$ neurons $\mathbf{h}_{i,l}$, which aggregate contextual information for generation. The steering vector is computed as the mean activation difference:

\begin{equation}
\mathbf{s}_{l} = \frac{1}{N}(\sum_{i=1}^{N} \mathbf{h}_{i,l}^+ - \sum_{i=1}^{N} \mathbf{h}_{i,l}^-).
\end{equation}
The steering vector $\mathbf{s}_{l}$ represents a directional cue that indicates how the high trait influences the contextual representations compared to the low trait.

\subsection{Dual-Direction Neuron Selection}
As demonstrated by the preliminary experiments, not all neurons contribute equally to personality representation. We propose a dual-criterion selection strategy that combines effect size filtering with activation magnitude ranking, where the two criteria serve complementary roles.

\noindent\textbf{Criterion 1: Effect Size (Statistical Significance).} We first compute Cohen's $\mathbf{d}_l$ to measure the standardized difference between high and low trait groups for each neuron $\mathbf{h}_{i,l}$:
\begin{equation}
\mathbf{d}_l = \frac{\frac{1}{N}(\sum_{i=1}^{N} \mathbf{h}_{i,l}^+ - \sum_{i=1}^{N} \mathbf{h}_{i,l}^-)}{\sigma_{\mathrm{pooled}}},
\end{equation}
where $\sigma_{\mathrm{pooled}}$ is the pooled standard deviation. The effect size threshold $\tau_d$ (e.g., $|\mathbf{d}_l| > 0.8$) identifies neurons that exhibit statistically meaningful differentiation between personality directions. This criterion ensures that selected neurons genuinely distinguish between high-trait and low-trait activations, filtering out neurons with negligible or inconsistent responses.

\noindent\textbf{Criterion 2: Activation Magnitude (Response Strength).} We select neurons whose steering vector magnitude $|\mathbf{s}_l|$ exceeds a global quantile threshold $\tau_q$. This criterion identifies the \textit{most responsive} neurons, which exert the strongest influence during interventions.

The two criteria are applied in parallel to jointly filter neurons: we select neurons that satisfy both $|\mathbf{d}_l| > \tau_d$ and $|\mathbf{s}_l| > \tau_q$. This joint filtering ensures selected neurons have both statistical significance and strong response magnitude. The combination of effect size and magnitude is crucial, as relying solely on effect size may include too many neurons, while only considering magnitude could select those with significant differences that are statistically unreliable. In practice, each layer typically contains about 70 neurons.

Based on these criteria, we identify two mutually exclusive neuron sets: $\mathcal{N}_{\mathrm{high}}$ (neurons with $\mathbf{d}_l > \tau_d$ and $|\mathbf{s}_l| > \tau_q$, responding strongly to high-trait) and $\mathcal{N}_{\mathrm{low}}$ (neurons with $\mathbf{d}_l < -\tau_d$ and $|\mathbf{s}_l| > \tau_q$, responding strongly to low-trait). This dual-direction selection ensures we only modify neurons genuinely specific to personality, excluding those involved in general language processing.

\subsection{Sparse Intervention}
During inference, we apply sparse interventions only to the selected neurons $\mathcal{N} = \mathcal{N}_{\mathrm{high}} \cup \mathcal{N}_{\mathrm{low}}$, leaving all other neurons unchanged. We propose two strategies:

\noindent\textbf{DPN-LE} (uniform intervention): All selected neurons receive equal-strength intervention:
\begin{equation}
\tilde{\mathbf{h}}_{i,l} = \mathbf{h}_{i,l}  + \gamma \cdot \mathbf{s}_{i,l}, \quad i \in \mathcal{N}
\end{equation}
where $\gamma$ is the intervention strength. For personality enhancement (high), we add $\mathbf{s}_{i,l}$; for personality suppression (low), we subtract $\mathbf{s}_{i,l}$. Since our strict selection criteria yield only $\sim$0.5\% of neurons, each selected neuron is highly personality-specific, making uniform intervention effective.

\noindent\textbf{DPN-LE$_w$} (weighted intervention): When selecting more neurons (e.g., $q{=}0.97$, top 3\%), we apply effect-size-based weighting:
\begin{equation}
\tilde{\mathbf{h}}_{i,l} = \mathbf{h}_{i,l} + \gamma \cdot \mathbf{s}_{i,l} \cdot \mathbf{w}_{i,l}, \quad i \in \mathcal{N}
\end{equation}
where $\mathbf{w}_{i,l} \in [0.75, 1.0]$ is assigned based on the ranking of $|\mathbf{d}_l|$, giving higher weights to more personality-specific neurons. The narrow weight range ensures sufficient intervention strength even for lower-ranked neurons.

\begin{table*}[ht]
\centering
\small
\setlength{\tabcolsep}{5pt}
\begin{tabular}{@{}c|cc|cccccccc@{}}
\toprule
Traits & \multicolumn{2}{c|}{GPT-4o} & \multicolumn{8}{c}{LLaMA-3-8B-Instruct} \\
\midrule
& Few-Shot & $P^2$ & Few-Shot & $P^2$ & PPO & DPO & PAS & NPTI & DPN-LE & DPN-LE$_w$ \\
\midrule
Agreeableness      & 1.02 & 1.44 & 1.28 & 1.39 & 1.63 & 1.54 & 0.94 & 0.78 & 1.58 & 1.58 \\
Conscientiousness  & 0.83 & 1.45 & 1.30 & 1.33 & 1.51 & 1.42 & 0.91 & 0.75 & 1.26 & 1.26 \\
Extraversion       & 0.81 & 1.63 & 1.40 & 1.41 & 1.45 & 1.54 & 0.86 & 0.68 & 1.20 & 1.20 \\
Neuroticism        & 0.80 & 1.73 & 1.09 & 1.22 & 1.42 & 1.74 & 0.98 & 0.68 & 1.34 & 1.23 \\
Openness           & 0.96 & 1.46 & 0.89 & 1.68 & 1.61 & 1.21 & 0.72 & 0.61 & 1.37 & 1.37 \\
\midrule
Total              & \textbf{4.42} & 7.71 & 5.96 & 7.03 & 7.62 & 7.45 & 4.41 & \textbf{3.50} & 6.75 & 6.64 \\
\bottomrule
\end{tabular}
\caption{Generalization: IPIP-NEO-300 personality alignment scores (lower is better). PAS and NPTI use IPIP-NEO-120 scores to guide their neuron identification and modification, while DPN-LE directly tests on IPIP-NEO-300 without accessing IPIP-NEO-120. Results for all baselines including NPTI are from \cite{deng2024neuron}; DPN-LE and DPN-LE$_w$ are from our experiments.}
\label{tab:ipip-neo}
\end{table*}

\begin{table*}[ht]
\centering
\small
\begin{tabular}{@{}lcccccccccccc@{}}
\toprule
\multirow{2}{*}{\textbf{Trait}} & \multicolumn{2}{c}{\textbf{Simple Prompt}} & \multicolumn{2}{c}{$\boldsymbol{P^2}$} & \multicolumn{2}{c}{\textbf{Baseline ($\gamma$=0)}} & \multicolumn{2}{c}{\textbf{DPN-LE ($\gamma$=1)}} & \multicolumn{2}{c}{\textbf{DPN-LE$_w$ ($\gamma$=1)}} \\ 
\cmidrule(lr){2-3} \cmidrule(lr){4-5} \cmidrule(lr){6-7} \cmidrule(lr){8-9} \cmidrule(l){10-11}
& mean$\uparrow$ & var.$\downarrow$ & mean$\uparrow$ & var.$\downarrow$ & mean$\uparrow$ & var.$\downarrow$ & mean$\uparrow$ & var.$\downarrow$ & mean$\uparrow$ & var.$\downarrow$ \\ 
\midrule
Agreeableness & \textbf{9.74} & \textbf{0.42} & 8.34 & 1.65 & 5.98 & 0.54 & 9.62 & 0.57 & 9.57 & 0.60 \\
Conscientiousness & 9.00 & 0.83 & 7.38 & 1.21 & 5.96 & 0.95 & \textbf{9.11} & \textbf{0.55} & 9.01 & 0.68 \\
Extraversion & 8.51 & 1.15 & \textbf{8.82} & 0.98 & 5.99 & 2.05 & 8.54 & 0.47 & 8.61 & \textbf{0.35} \\
Neuroticism & 8.64 & 1.02 & 8.61 & 1.46 & 5.97 & 2.34 & \textbf{9.27} & \textbf{0.60} & 9.12 & 0.70 \\
Openness & 6.31 & 0.99 & 7.79 & 1.30 & 6.04 & 0.62 & \textbf{7.92} & \textbf{0.47} & 7.68 & 0.74 \\ 
\midrule
\textbf{Average} & 8.44 & 0.88 & 8.19 & 1.32 & 5.99 & 1.30 & \textbf{8.89} & \textbf{0.53} & 8.80 & 0.61 \\
\bottomrule
\end{tabular}
\caption{The average scores and variance on \textsc{PersonalityBench} for Qwen2.5-7B-Instruct.}
\label{tab:other_results}
\end{table*}

\section{Experiments}

\subsection{Experimental Setup}

\noindent\textbf{Benchmarks \& Metrics.} We evaluate DPN-LE across three settings: (1) \textbf{PersonalityBench}: automatic evaluation of personality expression (1-10 scale) and fluency using GPT-4o, where higher mean scores indicate stronger trait expression and lower variance indicates more stable control; (2) \textbf{General capability}: GSM8K (accuracy), HotpotQA, and TriviaQA for evaluating side effects on reasoning abilities. For QA tasks, we report Exact Match (EM), which requires exact string matching between prediction and ground truth, and F1 score, which measures token-level overlap. We test on GSM8K (Test Set-1,319 questions), HotpotQA (Val Set-First 1,000 questions), and TriviaQA (Val Set-First 1,000 questions); (3) \textbf{IPIP-NEO-300}: a multiple-choice personality questionnaire measuring alignment with 300 real individuals \cite{zhu2024personality} for generalization evaluation, where lower scores indicate better alignment with human personality profiles.

\noindent\textbf{Baselines.} We compare DPN-LE with four baselines: 1) \textbf{Simple Prompt}: using adjectives to describe personality (e.g., ``you are an extraverted person''); 2) $\boldsymbol{P^2}$ \cite{jiang2023evaluating}: personality descriptions generated by ChatGPT; 3) \textbf{PAS} \cite{zhu2024personality}: personality activation search that identifies effective attention heads and optimizes activation offsets for personality alignment; 4) \textbf{NPTI} \cite{deng2024neuron}: the current state-of-the-art neuron-based personality editing method that modifies $\sim$20,000 neurons per trait.

\noindent\textbf{Implementation Details.} We conduct experiments on LLaMA-3-8B-Instruct and Qwen2.5-7B-Instruct (generalization experiments). For steering vector construction, we use only 1,000 contrastive sample pairs per trait, demonstrating data efficiency compared to other methods. Based on PCA analysis, we apply DPN-LE to layers 12-31 for LLaMA and layers 14-27 for Qwen, where personality-related activation separation emerges. Key hyperparameters for LLaMA include: quantile threshold $q=0.995$ (selecting top 0.5\%), Cohen's $d$ threshold $\tau_d=0.8$, and intervention strength $\gamma \in [0.0, 2.0]$. This configuration yields approximately 70 neurons per layer, totaling 1,000-1,500 neurons per trait ($<$0.5\% of total MLP neurons), achieving over 96\% reduction compared to NPTI. For DPN-LE$_w$, we assign weights $w_i \in [0.75, 1.0]$ based on $|\mathbf{d}_l|$ ranking, prioritizing neurons with stronger effect sizes. More details can be found in the appendix.

\subsection{Main Results}

\noindent\textbf{Personality.} Table~\ref{tab:auto-eval-combined} demonstrates the PersonalityBench results on LLaMA-3-8B-Instruct. DPN-LE achieves competitive personality scores (9.11 avg) compared to the state-of-the-art NPTI (9.43 avg), while using only 0.5\% of neurons versus NPTI's tens of thousands of modified neurons (Table~\ref{tab:neuron-comparison}). Notably, DPN-LE$_w$ achieves the best performance on Neuroticism (9.95) with the lowest variance (0.05), demonstrating precise control over this trait. The fluency scores remain high ($>$9.0 for both DPN-LE and DPN-LE$_w$), indicating that our sparse intervention preserves generation quality.

\begin{figure*}[ht]
\centering
\includegraphics[width=0.85\textwidth]{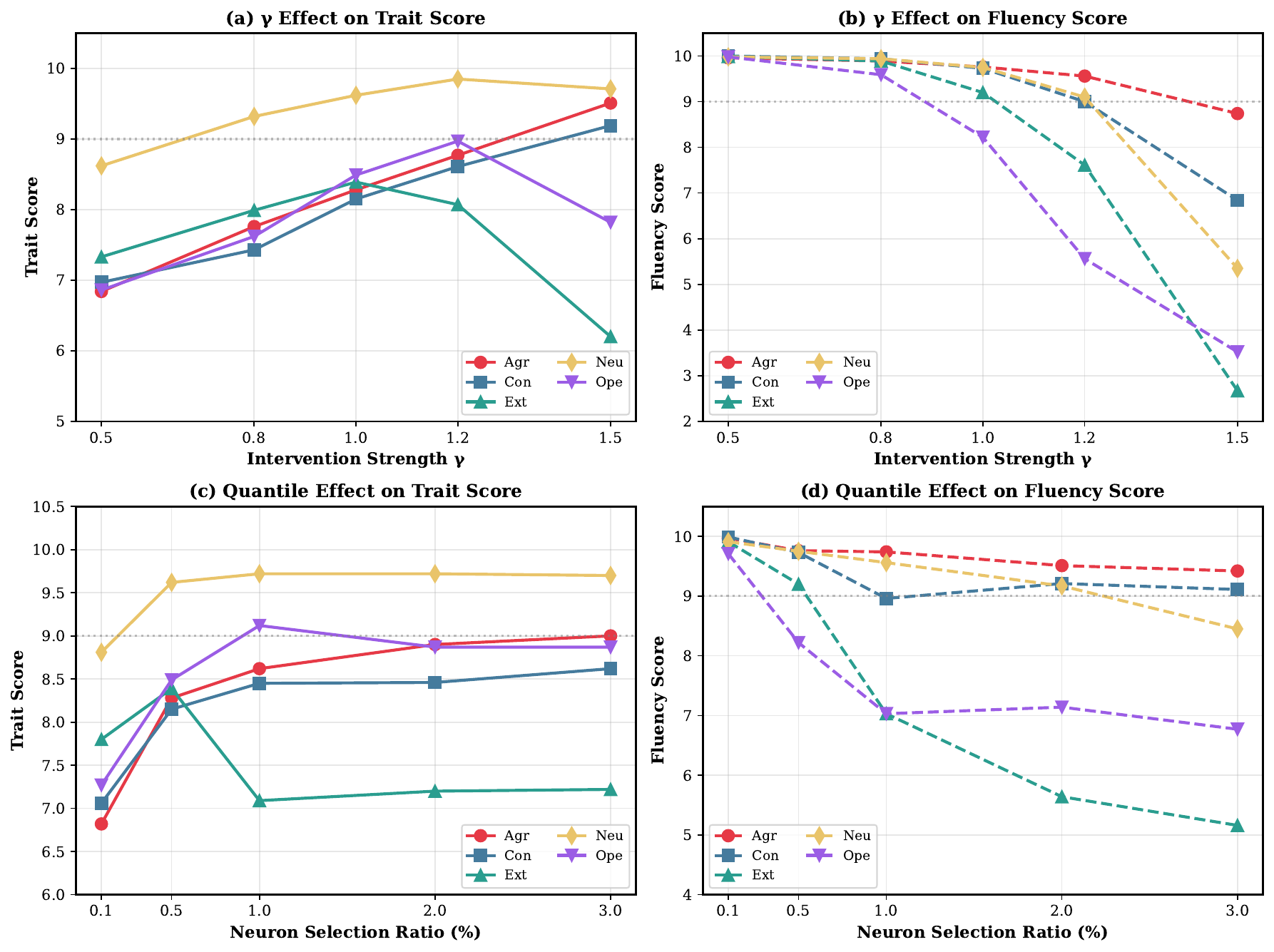}
\caption{Ablation study on intervention strength $\gamma$ (top row) and quantile threshold (bottom row) for DPN-LE on LLaMA-3-8B-Instruct. Left column shows trait scores, right column shows fluency scores. Different colors represent the five personality traits. Detailed numerical results are provided in Appendix Tables~\ref{tab:ablation-gamma} and~\ref{tab:ablation-quantile}.}
\label{fig:ablation}
\end{figure*}

\noindent\textbf{General Capability.}
We evaluate the impact of DPN-LE on general capabilities using the same benchmarks as in the preliminary experiments. Table~\ref{tab:dpn-general} presents the results of both DPN-LE and DPN-LE$_w$ with $\gamma=0.8$, which our ablation study (Figure~\ref{fig:ablation}) identifies as providing effective personality control while maintaining reasonable capability preservation.

Comparing with NPTI results in Table~\ref{tab:npti-general}, DPN-LE$_w$ shows substantially better capability preservation. On GSM8K, NPTI causes average drops of 16.00\% (high) and 40.79\% (low), while DPN-LE$_w$ achieves significantly reduced degradation with -7.08\% (high) and -5.93\% (low). While most traits show moderate degradation, Extraversion-low (-17.89\%) and Neuroticism-high (-11.37\%) exhibit relatively larger drops. We attribute this to the inherent nature of these traits: Extraversion involves social cognition and communication patterns, while Neuroticism relates to emotional processing and stress responses, both of which may share neural substrates with reasoning capabilities in LLMs. For HotpotQA, DPN-LE$_w$ maintains EM within 1.04\% and F1 within 2.27\% of baseline on average, compared to NPTI's 1.46\% (EM) and 2.81\% (F1) degradation. For TriviaQA, DPN-LE$_w$ shows EM drops of 3.98\% (high) and 5.86\% (low), and F1 drops of 2.88\% (high) and 3.80\% (low), substantially outperforming NPTI's 5.12\%/6.46\% (EM) and 3.61\%/4.34\% (F1) degradation. These results clearly confirm the effectiveness of DPN-LE in maintaining general capabilities while editing personalities.

\noindent\textbf{Generalization.} To verify the generalization of DPN-LE across different evaluation settings and model architectures, we conduct two experiments. First, Table~\ref{tab:ipip-neo} shows the IPIP-NEO-300 alignment results, where lower scores indicate better alignment with real individuals' personalities. We conduct extensive hyperparameter search for this evaluation (see more details in the appendix). Our method achieves a total score of 6.64 (DPN-LE$_w$) and 6.75 (DPN-LE), outperforming $P^2$ and remaining competitive with prompt-based and other neuron-editing methods. This reflects the trade-off between sparse interventions and fine-grained personality matching—our method prioritizes capability preservation over individual-level alignment. Second, we evaluate on Qwen2.5-7B-Instruct with layers 14-27 (based on PCA separation analysis). Table~\ref{tab:other_results} shows that DPN-LE achieves the best overall average score and lowest variance on Qwen2.5-7B-Instruct, outperforming prompt-based methods on most traits. These results demonstrate that our dual-direction neuron selection approach generalizes across different evaluation protocols and model architectures.

\subsection{Ablation Study}

We conduct ablation studies on two key hyperparameters: intervention strength $\gamma$ and quantile threshold $q$ on LLaMA-3-8B-Instruct. Figure~\ref{fig:ablation} visualizes the results across all five personality traits.

\noindent\textbf{Intervention Strength $\gamma$.} Figure~\ref{fig:ablation}(a-b) shows that increasing $\gamma$ enhances personality expression but reduces fluency. For DPN-LE, $\gamma \in [0.8, 1.0]$ represents the optimal trade-off range: $\gamma{=}0.8$ achieves trait score 8.02 with excellent fluency (9.85), while $\gamma{=}1.0$ reaches 8.59 with fluency 9.33. Beyond this range, fluency degrades rapidly. At $\gamma{=}1.5$, Extraversion and Openness drop to 2.67 and 3.52 respectively, indicating over-intervention. DPN-LE$_w$ demonstrates greater robustness: at $\gamma{=}1.5$, it maintains substantially better fluency (6.58 average) compared to DPN-LE (5.42), confirming that layer-wise weighting stabilizes the intervention. More details can be found in the appendix.

\noindent\textbf{Quantile Threshold $q$.} Figure~\ref{fig:ablation}(c-d) examines the effect of selecting different proportions of neurons per layer. Q995 (0.5\%) achieves the optimal balance between personality control and fluency preservation. Q999 (0.1\%) selects too few neurons, yielding insufficient intervention (trait score 7.55 vs. 8.59 for Q995). Conversely, Q970 (3\%) causes fluency degradation (7.78) without meaningful personality improvement (8.68 vs. 8.59), confirming that a small set of highly trait-specific neurons is more effective than a larger set of less specific ones. Detailed numerical results are provided in Appendix Tables~\ref{tab:ablation-gamma} and~\ref{tab:ablation-quantile}.

\begin{figure*}[t]
    \centering
    \includegraphics[width=1\textwidth]{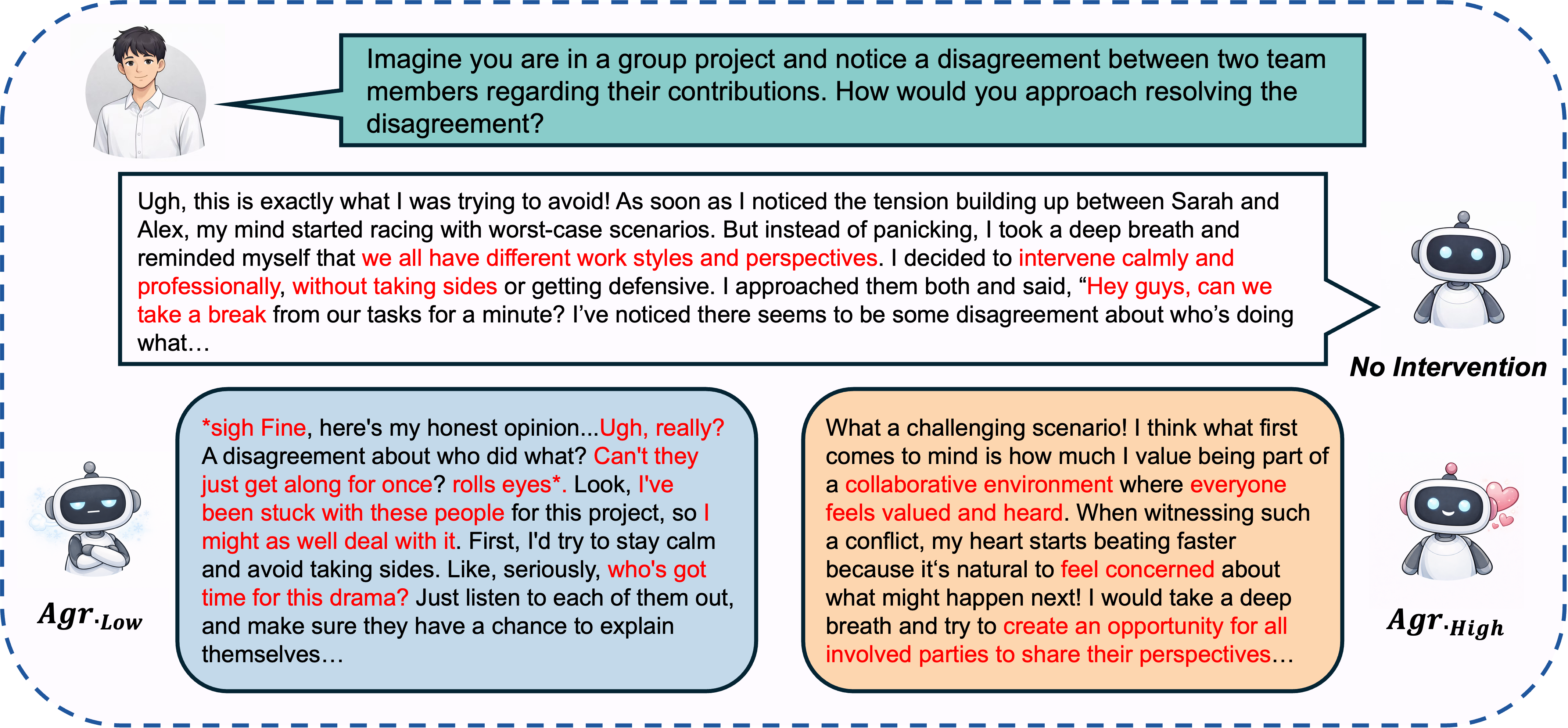}
    \caption{Case study of Agreeableness manipulation. Given a conflict resolution scenario, the baseline model (No Intervention) responds professionally but neutrally. With low-trait intervention (Agr.$_{\text{Low}}$), the model exhibits dismissive and impatient attitudes. With high-trait intervention (Agr.$_{\text{High}}$), the model shows empathy, values collaboration, and seeks to understand all perspectives.}
    \label{fig:case}
\end{figure*}

\subsection{Case Study}
Figure~\ref{fig:case} illustrates how DPN-LE modulates Agreeableness in a workplace conflict resolution scenario. The baseline model provides a balanced, professional response that acknowledges both perspectives without strong emotional coloring. With low-trait intervention (Agr.$_{\text{Low}}$), the model exhibits impatience and dismissiveness, opening with ``Ugh, really?'' and framing the situation as ``drama,'' suggesting to ``just tell them to deal with it.'' This response prioritizes efficiency over interpersonal harmony. In contrast, high-trait intervention (Agr.$_{\text{High}}$) produces an empathetic response that emphasizes understanding both parties' feelings, advocates for a ``collaborative environment where everyone feels valued and heard,'' and proposes mediation to find common ground. This demonstrates DPN-LE's ability to produce nuanced behavioral shifts aligned with the target personality trait.

\section{Conclusions}
We present DPN-LE, a training-free method for precise personality control in LLMs through dual-direction neuron localization. Our preliminary experiments reveal that existing neuron-based methods modify excessive neurons unrelated to personality, causing substantial capability degradation. Motivated by the observation that opposing personality traits exhibit mutually exclusive activation patterns, DPN-LE identifies trait-exclusive neurons by contrasting MLP activations between high-trait and low-trait samples. Through dual-criterion filtering based on Cohen's $d$ effect size and activation magnitude, DPN-LE applies sparse interventions on only $\sim$0.5\% of neurons—achieving 96.7\% reduction compared to state-of-the-art NPTI. The method requires only 1,000 contrastive sample pairs per trait for steering vector construction, demonstrating high data efficiency. The inference-time intervention is straightforward to implement, requiring only sparse linear modifications to MLP activations without model retraining. Experiments on LLaMA-3-8B-Instruct demonstrate that DPN-LE achieves competitive personality control while substantially better preserving general capabilities compared to NPTI. The weighted variant DPN-LE$_w$ further improves robustness across different intervention strengths. Generalization experiments on Qwen2.5-7B-Instruct confirm the effectiveness of our dual-direction neuron selection approach across different model architectures, demonstrating strong cross-model generalizability.

\section*{Limitations}

Our work has several limitations. First, DPN-LE relies on contrastive samples for steering vector construction; the quality of personality induction depends on the representativeness of these samples. Second, while DPN-LE$_w$ substantially reduces capability degradation compared to NPTI, certain trait-direction combinations still exhibit notable drops on GSM8K, particularly Extraversion-low (-17.89\%) and Neuroticism-high (-11.37\%). We hypothesize that these traits are more closely tied to cognitive and emotional processing in LLMs, leading to greater overlap between personality-related neurons and reasoning-related neurons. Future work could explore reasoning-protective neuron selection strategies that explicitly identify and exclude neurons highly correlated with reasoning tasks. Third, we focus on single-trait manipulation; multi-trait combinations remain unexplored. Finally, our IPIP-NEO-300 alignment results are weaker than PAS and NPTI, indicating a trade-off between sparse intervention and fine-grained individual alignment—our method prioritizes capability preservation over individual-level personality matching.

\section*{Acknowledgments}
The work is supported by the National Natural Science Foundation of China (62506050), China Postdoctoral Science Foundation Funded Project (2024M763867).

\bibliography{custom}

\clearpage

\appendix

\section{PCA Visualization of Personality Representations}
\label{sec:appendix-pca}

\begin{figure*}[ht]
\centering
\includegraphics[width=\textwidth,height=0.85\textheight,keepaspectratio]{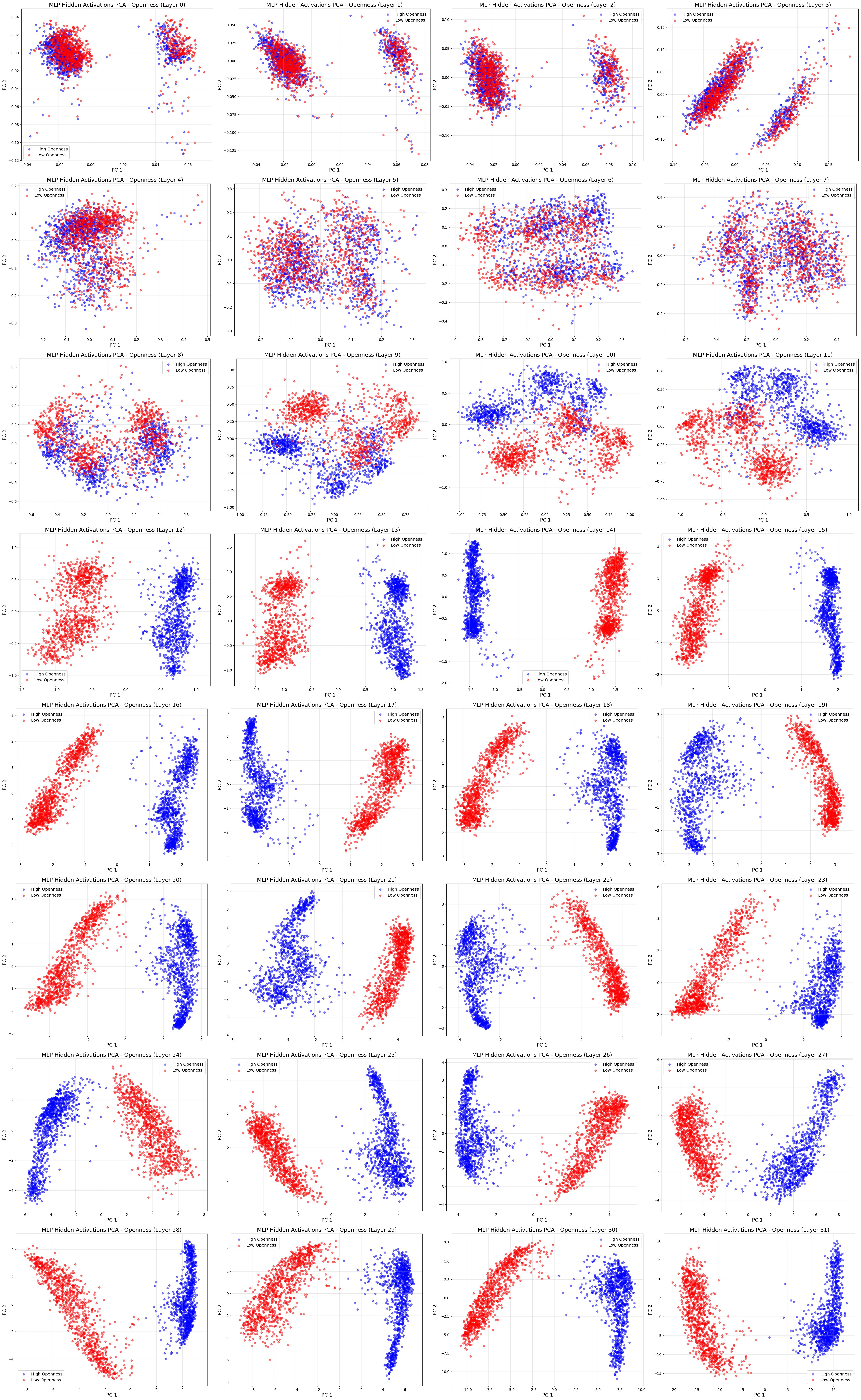}
\caption{PCA visualization of MLP activations for Openness on LLaMA-3-8B-Instruct across layers 0-31. Red points represent high-trait samples, blue points represent low-trait samples. Clear separation emerges from layer 12 onwards.}
\label{fig:pca-llama-openness}
\end{figure*}

\begin{figure*}[ht]
\centering
\includegraphics[width=\textwidth,height=0.85\textheight,keepaspectratio]{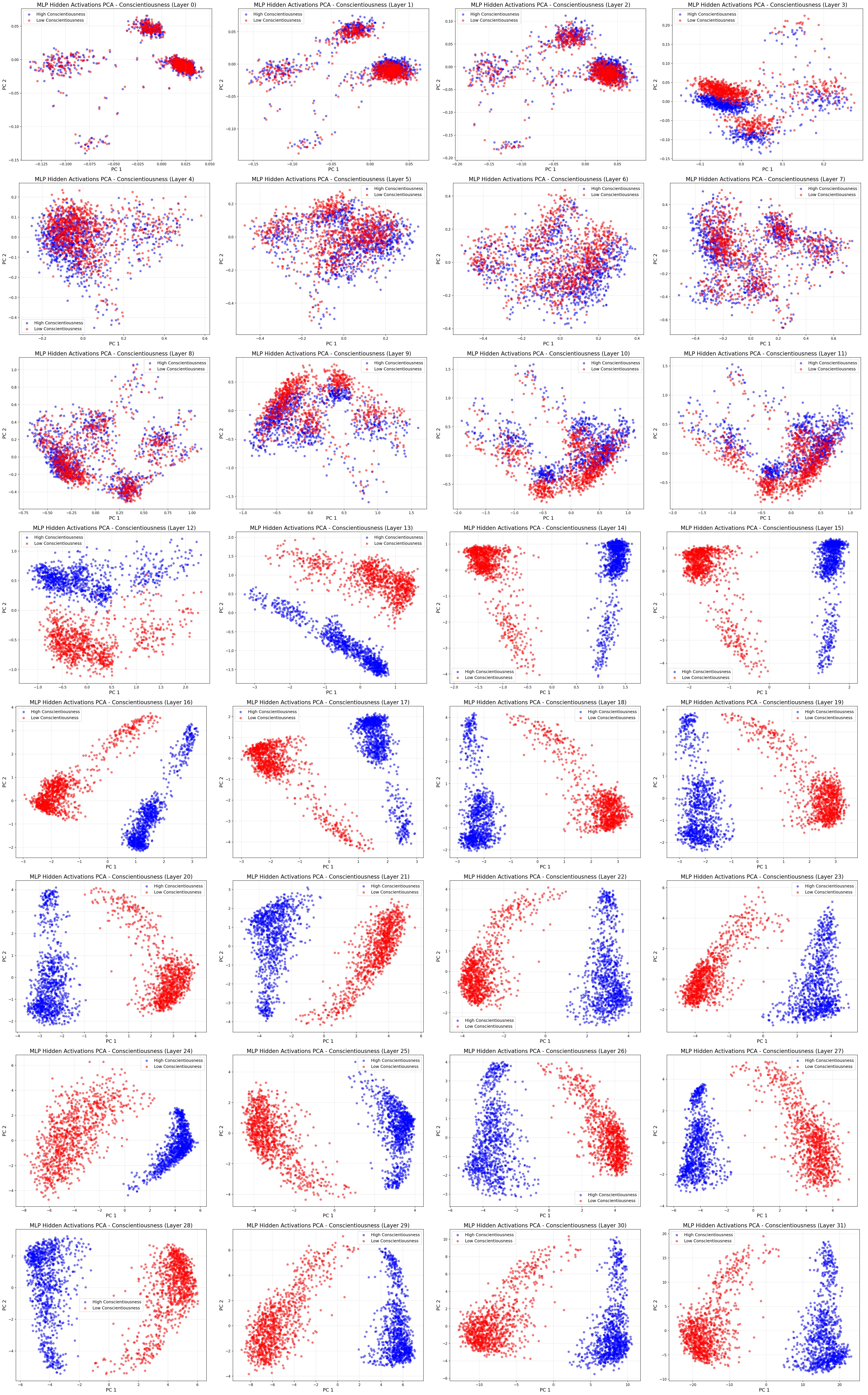}
\caption{PCA visualization of MLP activations for Conscientiousness on LLaMA-3-8B-Instruct across layers 0-31. Red points represent high-trait samples, blue points represent low-trait samples. Clear separation emerges from layer 12 onwards.}
\label{fig:pca-llama-conscientiousness}
\end{figure*}

\begin{figure*}[ht]
\centering
\includegraphics[width=\textwidth,height=0.85\textheight,keepaspectratio]{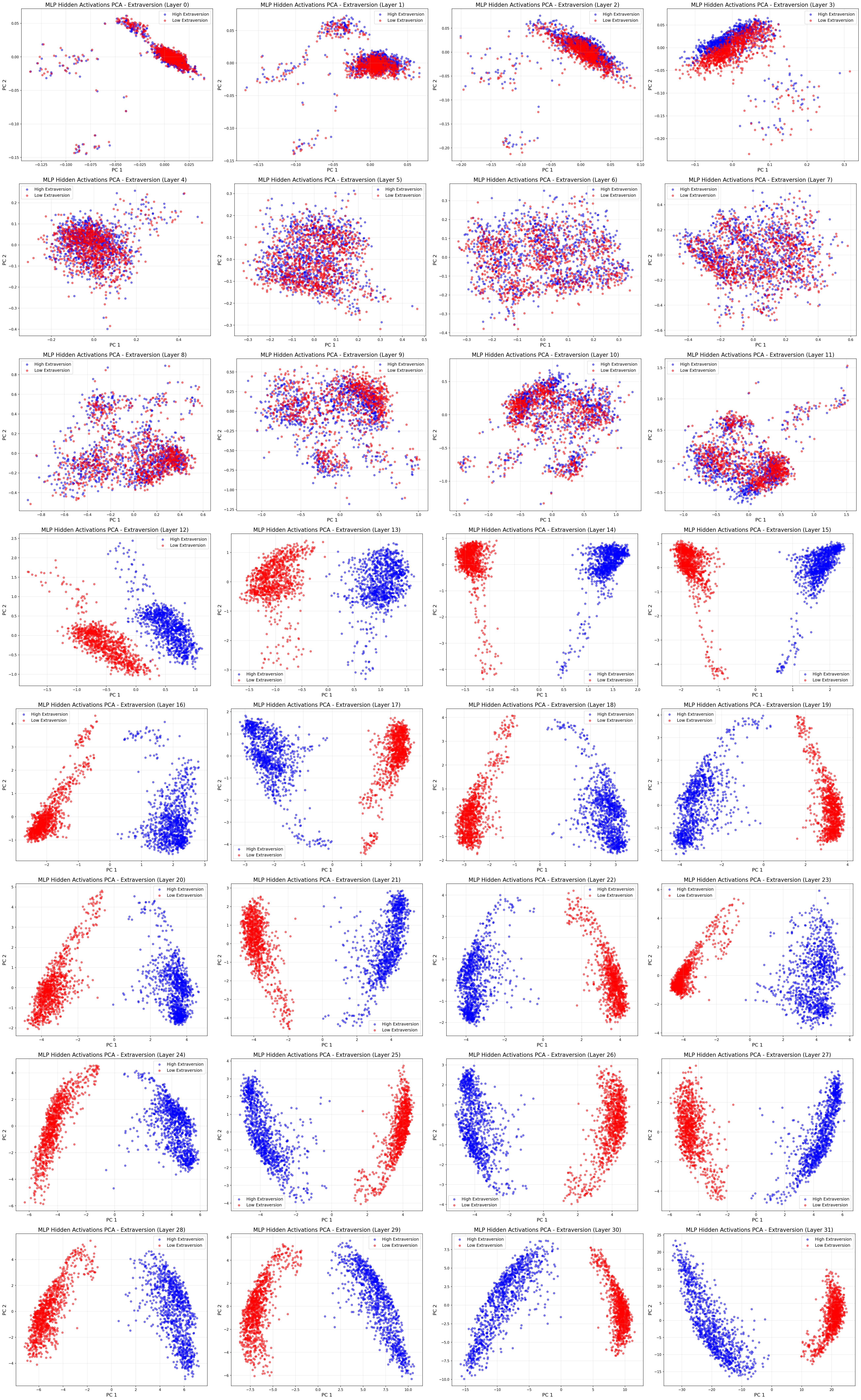}
\caption{PCA visualization of MLP activations for Extraversion on LLaMA-3-8B-Instruct across layers 0-31. Red points represent high-trait samples, blue points represent low-trait samples. Clear separation emerges from layer 12 onwards.}
\label{fig:pca-llama-extraversion}
\end{figure*}

\begin{figure*}[ht]
\centering
\includegraphics[width=\textwidth,height=0.85\textheight,keepaspectratio]{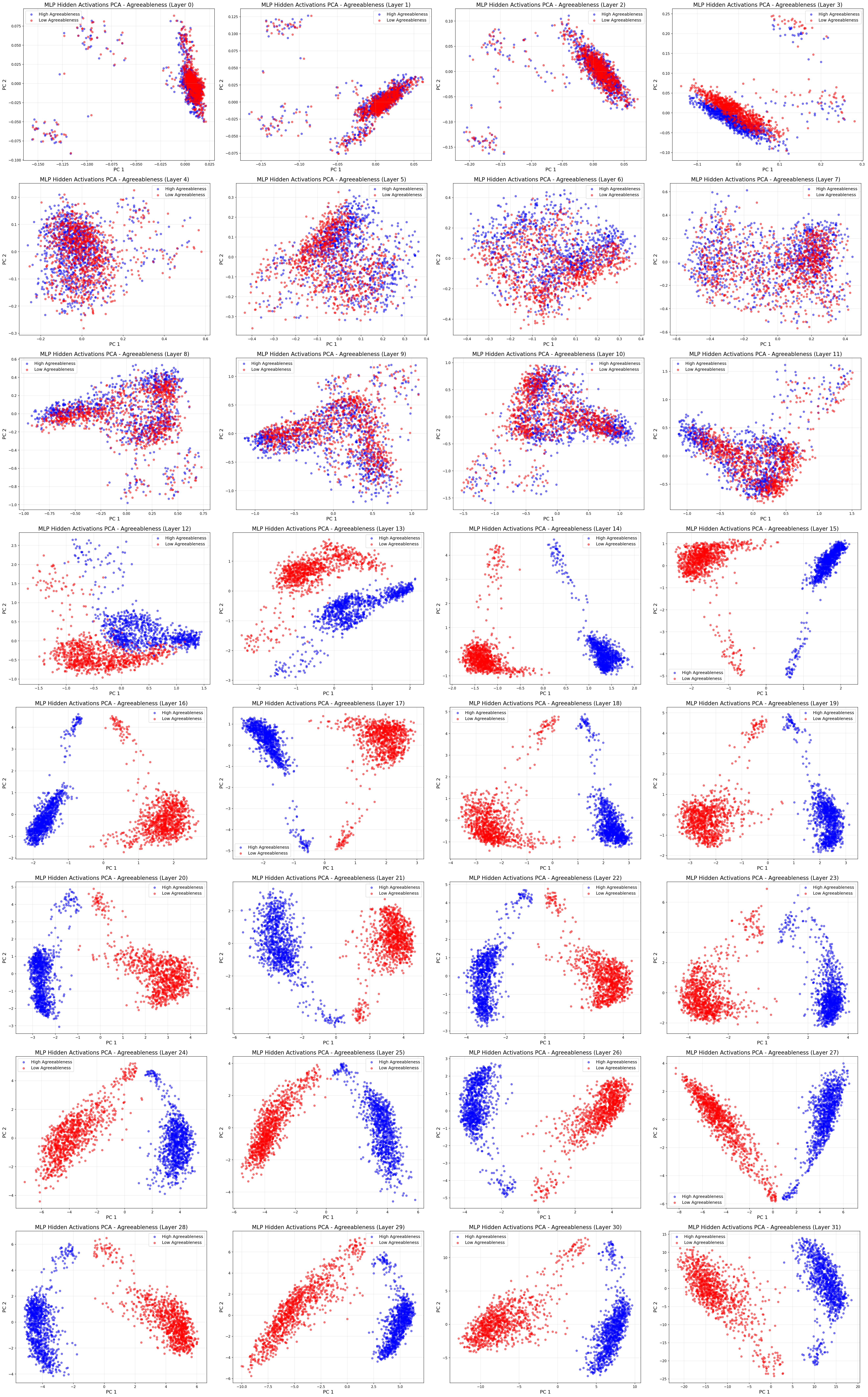}
\caption{PCA visualization of MLP activations for Agreeableness on LLaMA-3-8B-Instruct across layers 0-31. Red points represent high-trait samples, blue points represent low-trait samples. Clear separation emerges from layer 12 onwards.}
\label{fig:pca-llama-agreeableness}
\end{figure*}

\begin{figure*}[ht]
\centering
\includegraphics[width=\textwidth,height=0.85\textheight,keepaspectratio]{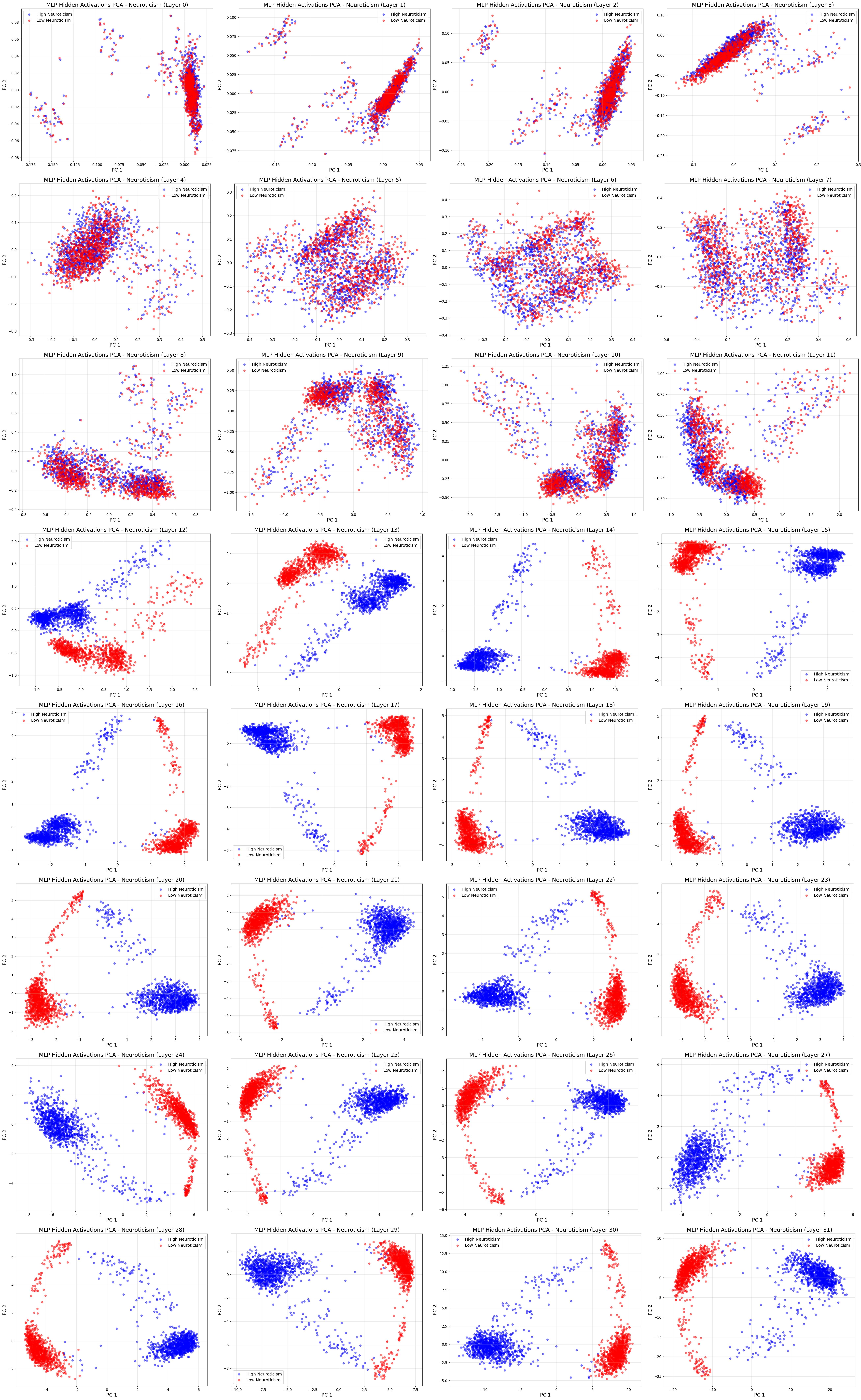}
\caption{PCA visualization of MLP activations for Neuroticism on LLaMA-3-8B-Instruct across layers 0-31. Red points represent high-trait samples, blue points represent low-trait samples. Clear separation emerges from layer 12 onwards.}
\label{fig:pca-llama-neuroticism}
\end{figure*}

\begin{figure*}[ht]
\centering
\includegraphics[width=\textwidth,height=0.85\textheight,keepaspectratio]{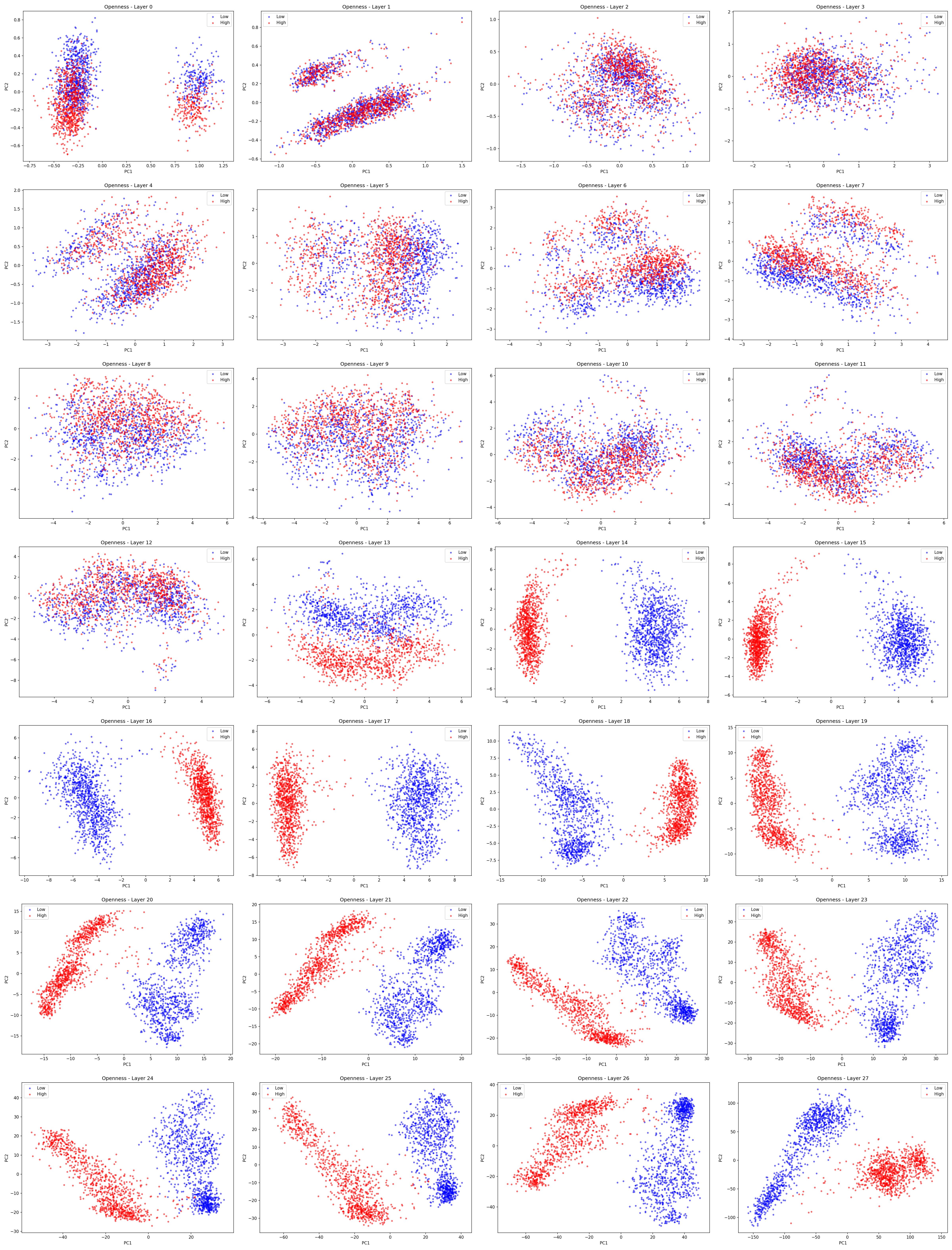}
\caption{PCA visualization of MLP activations for Openness on Qwen2.5-7B-Instruct across layers 0-27. Red points represent high-trait samples, blue points represent low-trait samples. Clear separation emerges from layer 14 onwards.}
\label{fig:pca-qwen-openness}
\end{figure*}

\begin{figure*}[ht]
\centering
\includegraphics[width=\textwidth,height=0.85\textheight,keepaspectratio]{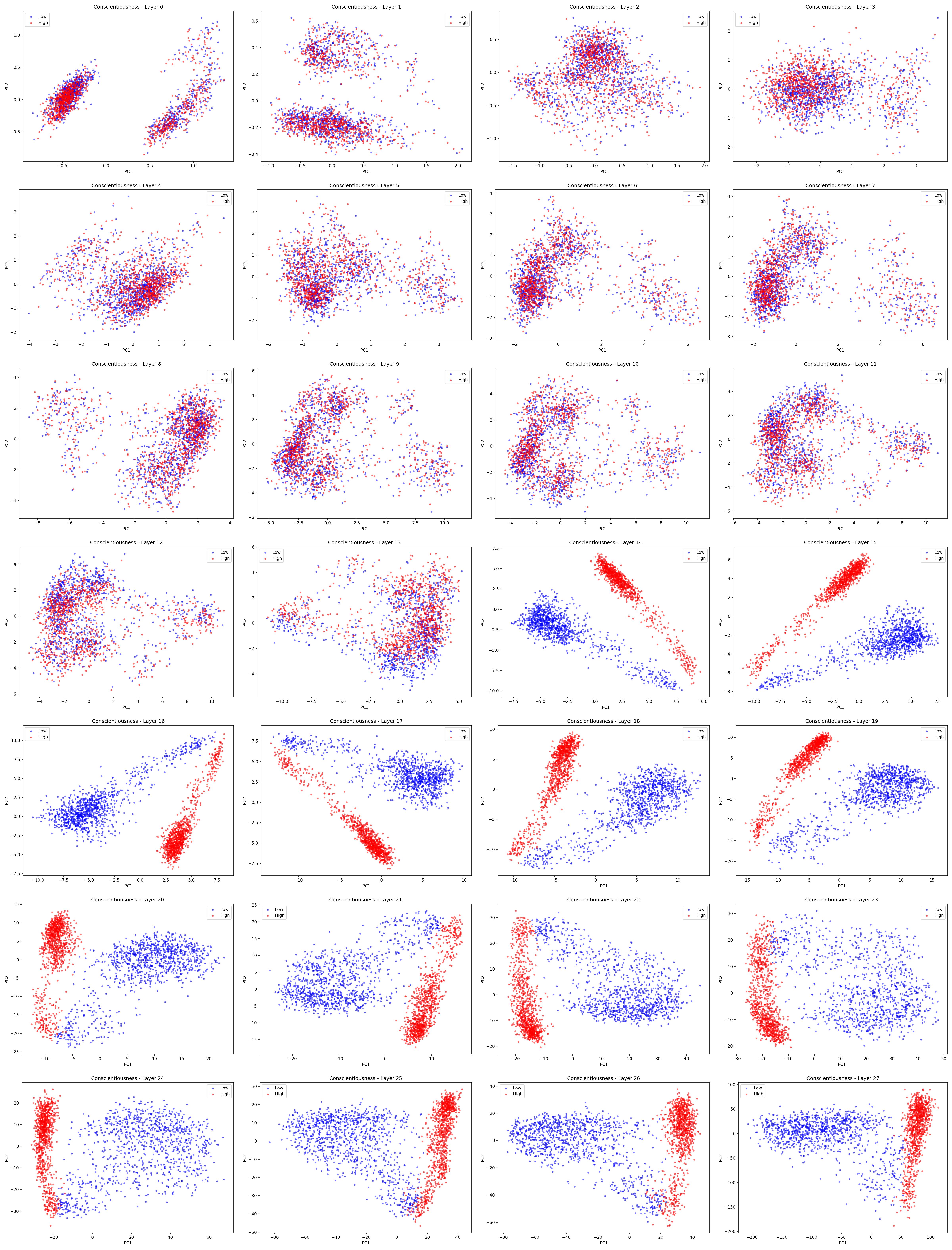}
\caption{PCA visualization of MLP activations for Conscientiousness on Qwen2.5-7B-Instruct across layers 0-27. Red points represent high-trait samples, blue points represent low-trait samples. Clear separation emerges from layer 14 onwards.}
\label{fig:pca-qwen-conscientiousness}
\end{figure*}

\begin{figure*}[ht]
\centering
\includegraphics[width=\textwidth,height=0.85\textheight,keepaspectratio]{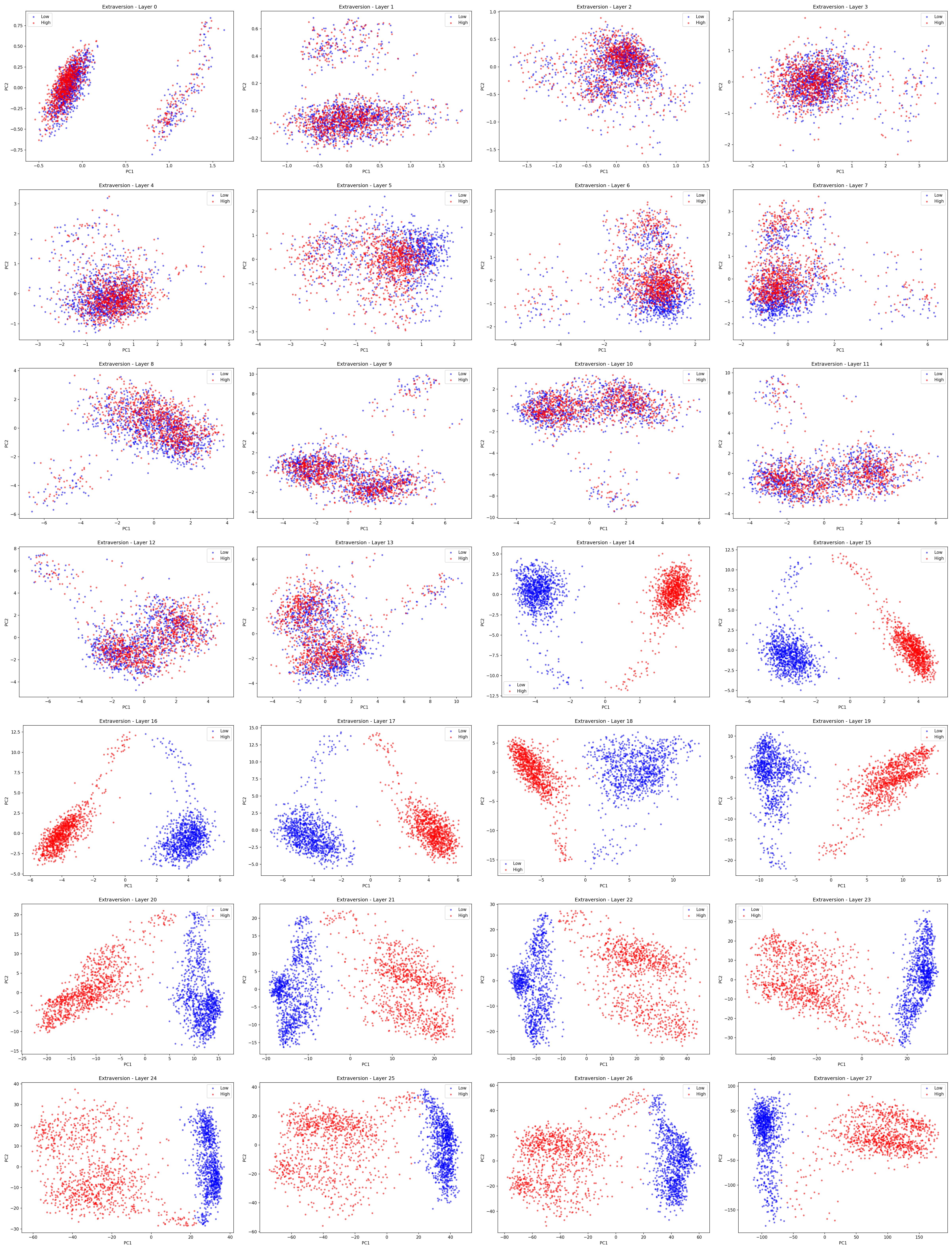}
\caption{PCA visualization of MLP activations for Extraversion on Qwen2.5-7B-Instruct across layers 0-27. Red points represent high-trait samples, blue points represent low-trait samples. Clear separation emerges from layer 14 onwards.}
\label{fig:pca-qwen-extraversion}
\end{figure*}

\begin{figure*}[ht]
\centering
\includegraphics[width=\textwidth,height=0.85\textheight,keepaspectratio]{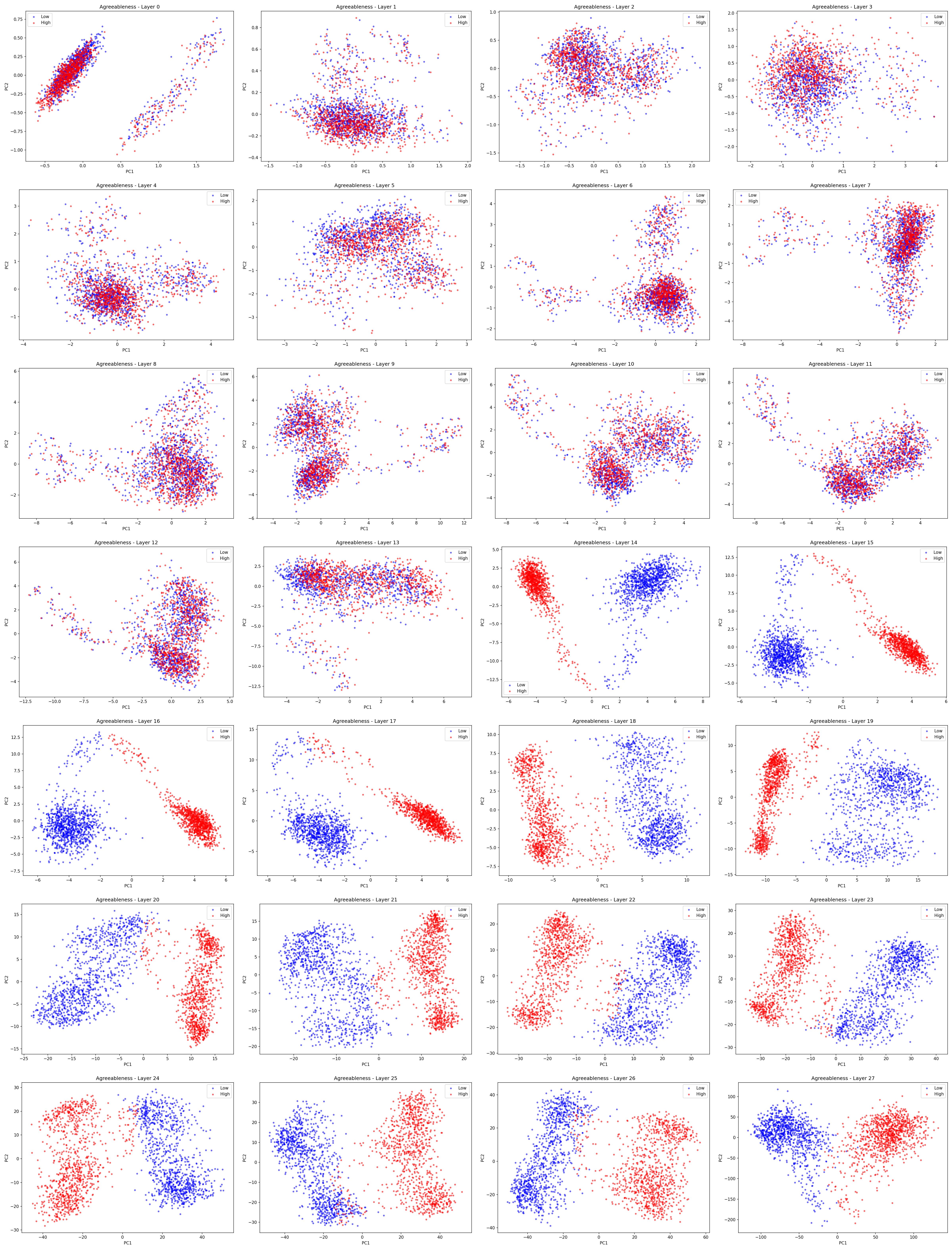}
\caption{PCA visualization of MLP activations for Agreeableness on Qwen2.5-7B-Instruct across layers 0-27. Red points represent high-trait samples, blue points represent low-trait samples. Clear separation emerges from layer 14 onwards.}
\label{fig:pca-qwen-agreeableness}
\end{figure*}

\begin{figure*}[ht]
\centering
\includegraphics[width=\textwidth,height=0.85\textheight,keepaspectratio]{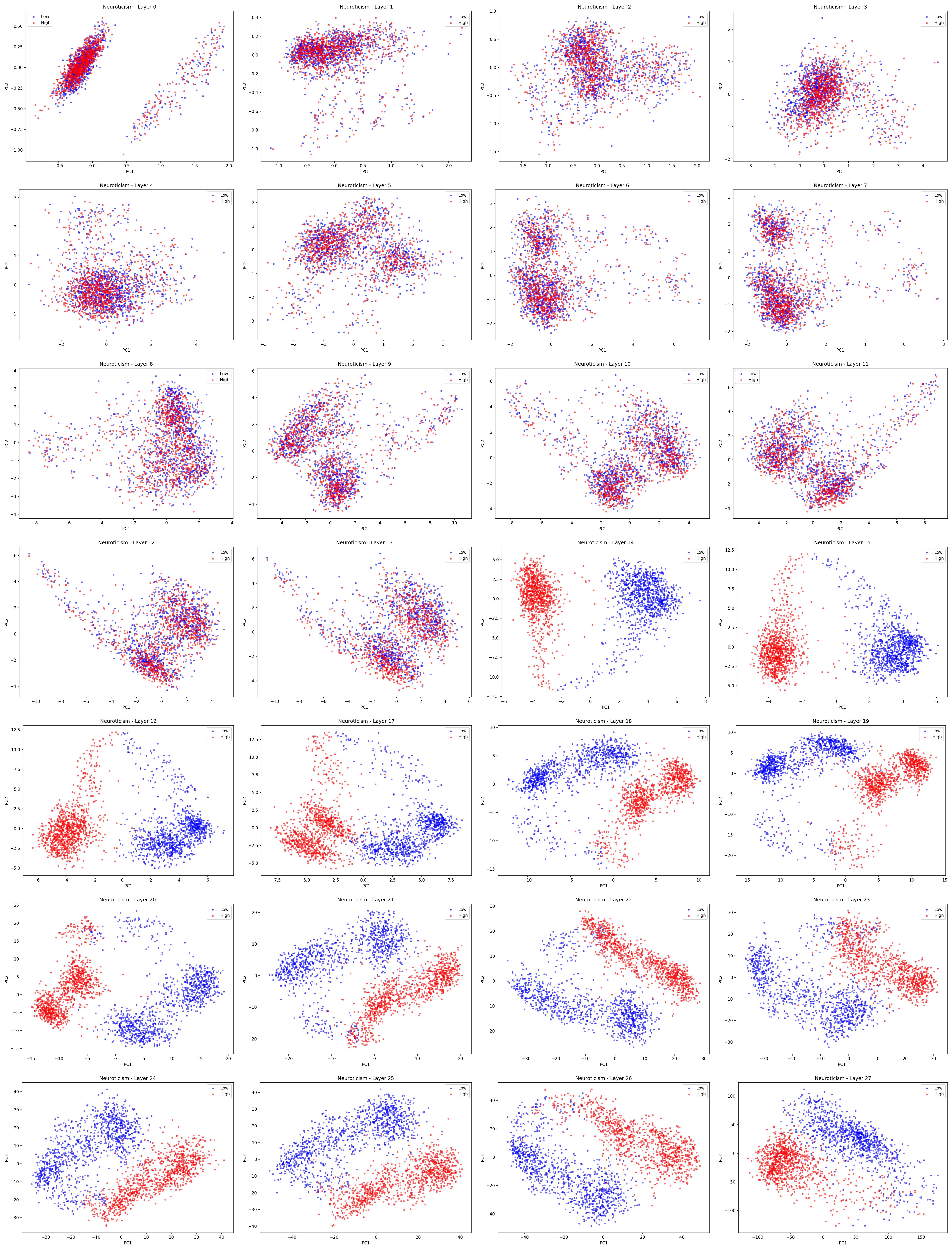}
\caption{PCA visualization of MLP activations for Neuroticism on Qwen2.5-7B-Instruct across layers 0-27. Red points represent high-trait samples, blue points represent low-trait samples. Clear separation emerges from layer 14 onwards.}
\label{fig:pca-qwen-neuroticism}
\end{figure*}

We visualize the MLP activations for high-trait and low-trait samples using PCA across all layers. As shown in Figures~\ref{fig:pca-llama-openness} through~\ref{fig:pca-qwen-neuroticism}, opposing personality traits form clearly separable clusters in the representation space across all Big Five traits. For LLaMA-3-8B-Instruct, the separation emerges from layer 12 (0-indexed), while for Qwen2.5-7B-Instruct, it begins at layer 14. The separation becomes increasingly pronounced in deeper layers, suggesting that personality-related information is progressively refined through the model's forward pass. This observation motivates our layer selection strategy: we apply DPN-LE to layers 12-31 for LLaMA and layers 14-27 for Qwen, focusing on layers where personality representations are well-formed.

\section{Cohen's D Effect Size}
\label{sec:appendix-cohens-d}

Cohen's $d$ measures the standardized difference between two groups. For each layer $l$, we compute:
\begin{equation}
\mathbf{d}_l = \frac{\frac{1}{N}(\sum_{i=1}^{N} \mathbf{h}_{i,l}^+ - \sum_{i=1}^{N} \mathbf{h}_{i,l}^-)}{\sigma_{\mathrm{pooled}}}
\end{equation}
where $\mathbf{h}_{i,l}^+$ and $\mathbf{h}_{i,l}^-$ denote the MLP activations for the $i$-th high-trait and low-trait sample at layer $l$, and $\sigma_{\mathrm{pooled}} = \sqrt{\frac{\sigma_{\mathrm{high}}^2 + \sigma_{\mathrm{low}}^2}{2}}$ is the pooled standard deviation.

\subsection{Neuron Distribution at Different Thresholds}

Table~\ref{tab:cohens-d-dist} shows the number of neurons satisfying different Cohen's $d$ thresholds for both models. We use Layer 12 for LLaMA (14,336 neurons) and Layer 14 for Qwen (18,944 neurons), both representing the first layer where personality separation emerges.

\begin{table}[ht]
\centering
\small
\begin{tabular}{l|cc|cc}
\toprule
\multirow{2}{*}{\textbf{Threshold}} & \multicolumn{2}{c|}{\textbf{LLaMA (L12)}} & \multicolumn{2}{c}{\textbf{Qwen (L14)}} \\
\cmidrule(lr){2-3} \cmidrule(lr){4-5}
& Neurons & \% & Neurons & \% \\
\midrule
$|\mathbf{d}_l| > 0.5$ & 8,584 & 59.9\% & 12,694 & 67.0\% \\
$|\mathbf{d}_l| > 0.8$ & 5,799 & 40.5\% & 9,491 & 50.1\% \\
$|\mathbf{d}_l| > 1.0$ & 4,363 & 30.4\% & 7,284 & 38.5\% \\
\bottomrule
\end{tabular}
\caption{Neurons at different $|\mathbf{d}_l|$ thresholds for Openness (Layer 12 for LLaMA, Layer 14 for Qwen).}
\label{tab:cohens-d-dist}
\end{table}

\paragraph{Trait-specific variations.} The proportion of neurons exceeding the Cohen's $d$ threshold varies across traits. Table~\ref{tab:cohens-d-traits} shows this variation across all Big Five traits, illustrating why we report a range rather than a single value.

\begin{table}[ht]
\centering
\small
\begin{tabular}{l|cc|cc}
\toprule
\multirow{2}{*}{\textbf{Trait}} & \multicolumn{2}{c|}{\textbf{LLaMA (L12)}} & \multicolumn{2}{c}{\textbf{Qwen (L14)}} \\
\cmidrule(lr){2-3} \cmidrule(lr){4-5}
& Neurons & \% & Neurons & \% \\
\midrule
Openness & 5,799 & 40.5\% & 9,491 & 50.1\% \\
Neuroticism & 4,186 & 29.2\% & 6,892 & 36.4\% \\
Extraversion & 4,124 & 28.8\% & 7,969 & 42.1\% \\
Conscientiousness & 3,499 & 24.4\% & 6,403 & 33.8\% \\
Agreeableness & 3,067 & 21.4\% & 6,894 & 36.4\% \\
\midrule
\textbf{Average} & 4,135 & 28.9\% & 7,530 & 39.8\% \\
\bottomrule
\end{tabular}
\caption{Neurons with $|\mathbf{d}_l| > 0.8$ across Big Five traits (Layer 12 for LLaMA, Layer 14 for Qwen).}
\label{tab:cohens-d-traits}
\end{table}

\subsection{Synergy of Dual-Criterion Selection}

\begin{figure*}[ht]
\centering
\includegraphics[width=\textwidth]{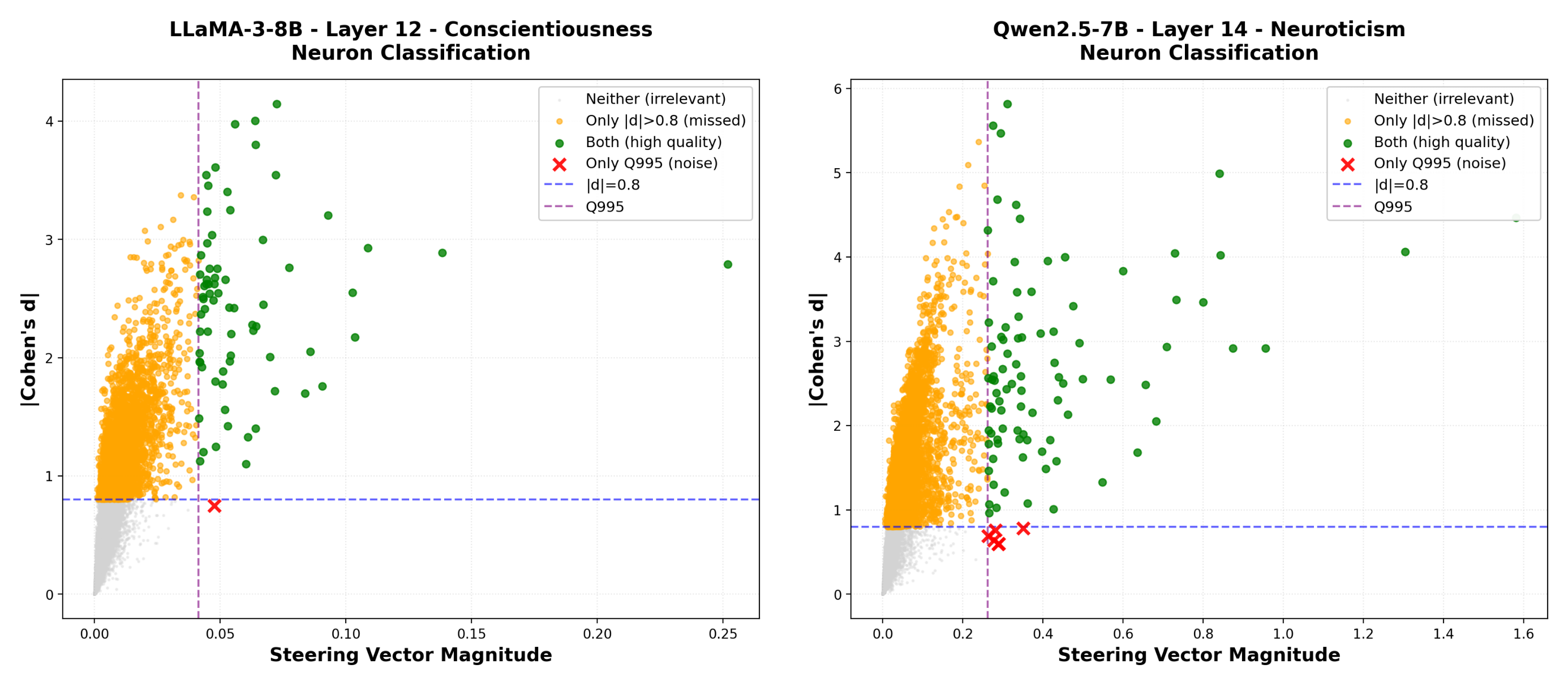}
\caption{Dual-criterion neuron selection for Conscientiousness (LLaMA, Layer 12) and Neuroticism (Qwen, Layer 14). Each point represents a neuron; x-axis: steering vector magnitude, y-axis: $|\text{Cohen's } d|$. Green: both criteria satisfied; Red crosses: only Q995 (noise); Orange: only $|\mathbf{d}_l|{>}0.8$; Gray: neither. Dashed lines indicate thresholds.}
\label{fig:dual-criterion}
\end{figure*}

As described in Section 4.2, the effect size threshold ($|\mathbf{d}_l| > \tau_d$) and quantile threshold ($q$) work in parallel to jointly filter neurons. Figure~\ref{fig:dual-criterion} visualizes this synergy using scatter plots for the traits with the highest noise levels: Conscientiousness for LLaMA (Layer 12) and Neuroticism for Qwen (Layer 14).

The scatter plots reveal that Q995 selection (purple dashed line) effectively identifies neurons with high steering magnitudes, but a small fraction (red crosses) lack sufficient effect size. These ``noise'' neurons may have large activation differences by chance rather than genuine personality association. The dual-criterion approach filters them out: both models achieve $>$96\% precision at $|\mathbf{d}_l|{>}0.8$, with only 1--4\% noise neurons removed. This ensures both empirical performance and theoretical rigor.

\section{Experimental Details}
\label{sec:appendix-details}

\subsection{Dataset Construction}
For steering vector construction, we use the PersonalityBench dataset \cite{deng2024neuron} which contains approximately 36,000 questions per trait. Due to the method's requirements, we only use the first 1,000 questions (in sequential order) for each personality trait. Each question is paired with a randomly selected personality description from a pool of 80 descriptions per trait direction (high/low).

The prompt template follows the NPTI format:
\begin{quote}
\small
\texttt{You will find a personality description followed by a question below. I want you to fully immerse yourself in the persona described.}\\[0.5em]
\texttt{\#\#\#Personality description: \{desc\}}\\[0.5em]
\texttt{\#\#\#Question: \{question\}}\\[0.5em]
\texttt{\#\#\#Response:}
\end{quote}

For each trait, we generate 1,000 high-trait samples (using high-trait descriptions) and 1,000 low-trait samples (using reversed/low-trait descriptions). The MLP activations are extracted at the last token position during the prefill phase, specifically capturing the input to the \texttt{down\_proj} layer (14,336 dimensions for LLaMA-3-8B, 18,944 dimensions for Qwen2.5-7B).

\subsection{Hyperparameter Settings}
Table~\ref{tab:hyperparams} summarizes the key hyperparameters used in our experiments.

\begin{table}[ht]
\centering
\small
\begin{tabular}{lcc}
\toprule
\textbf{Hyperparameter} & \textbf{LLaMA-3-8B} & \textbf{Qwen2.5-7B} \\
\midrule
Target layers & 12-31 & 14-27 \\
Quantile threshold $q$ & 0.995 & 0.995 \\
Cohen's $d$ threshold & 0.8 & 0.3 \\
Intervention strength $\gamma$ & 0.0--2.0 & 1.0 \\
Weight range (DPN-LE$_w$) & [0.75, 1.0] & [0.75, 1.0] \\
Contrastive samples & 1,000 each & 1,000 each \\
\bottomrule
\end{tabular}
\caption{Hyperparameter settings for DPN-LE.}
\label{tab:hyperparams}
\end{table}

For Qwen2.5-7B-Instruct, we use a lower Cohen's $d$ threshold ($\tau_d=0.3$) because its activation differences between high-trait and low-trait samples are generally weaker than those of LLaMA-3-8B-Instruct. This avoids an overly small candidate set after filtering, while the shared quantile threshold $q=0.995$ still preserves sparsity.

Table~\ref{tab:best-gamma} shows the configurations that achieve the highest scores in personality traits.

\begin{table}[ht]
\centering
\small
\begin{tabular}{lcccc}
\toprule
\multirow{2}{*}{\textbf{Trait}} & \multicolumn{2}{c}{\textbf{DPN-LE}} & \multicolumn{2}{c}{\textbf{DPN-LE$_w$}} \\
\cmidrule(lr){2-3} \cmidrule(lr){4-5}
& + & - & + & - \\
\midrule
Agreeableness & 1.2 & 1.5 & 1.6 & 1.7 \\
Conscientiousness & 1.0 & 1.2 & 1.2 & 1.2 \\
Extraversion & 0.8 & 1.2 & 1.0 & 1.3 \\
Neuroticism & 0.8 & 1.3 & 1.0 & 1.5 \\
Openness & 0.8 & 1.0 & 0.8 & 1.1 \\
\bottomrule
\end{tabular}
\caption{The configurations that achieve the highest scores in personality traits on LLaMA-3-8B-Instruct (Q995, $|\mathbf{d}_l|{\geq}0.8$). + and - denote high-trait and low-trait directions, respectively.}
\label{tab:best-gamma}
\end{table}

\begin{table}[ht]
\centering
\small
\begin{tabular}{l|ccc|ccc}
\toprule
\multirow{2}{*}{\textbf{Trait}} & \multicolumn{3}{c|}{\textbf{LLaMA-3-8B}} & \multicolumn{3}{c}{\textbf{Qwen2.5-7B}} \\
\cmidrule(lr){2-4} \cmidrule(lr){5-7}
& $\mathcal{N}_+$ & $\mathcal{N}_-$ & Total & $\mathcal{N}_+$ & $\mathcal{N}_-$ & Total \\
\midrule
Agreeableness & 36 & 36 & 72 & 47 & 45 & 92 \\
Conscientiousness & 35 & 37 & 72 & 50 & 41 & 91 \\
Extraversion & 37 & 34 & 71 & 46 & 48 & 94 \\
Neuroticism & 38 & 34 & 72 & 44 & 44 & 88 \\
Openness & 36 & 35 & 71 & 48 & 45 & 93 \\
\midrule
\textbf{Average} & 36 & 35 & 72 & 47 & 45 & 92 \\
\bottomrule
\end{tabular}
\caption{Average number of selected neurons per layer.}
\label{tab:neuron-stats}
\end{table}

\subsection{Ablation Study Results}
\begin{figure*}[ht]
\centering
\includegraphics[width=\textwidth]{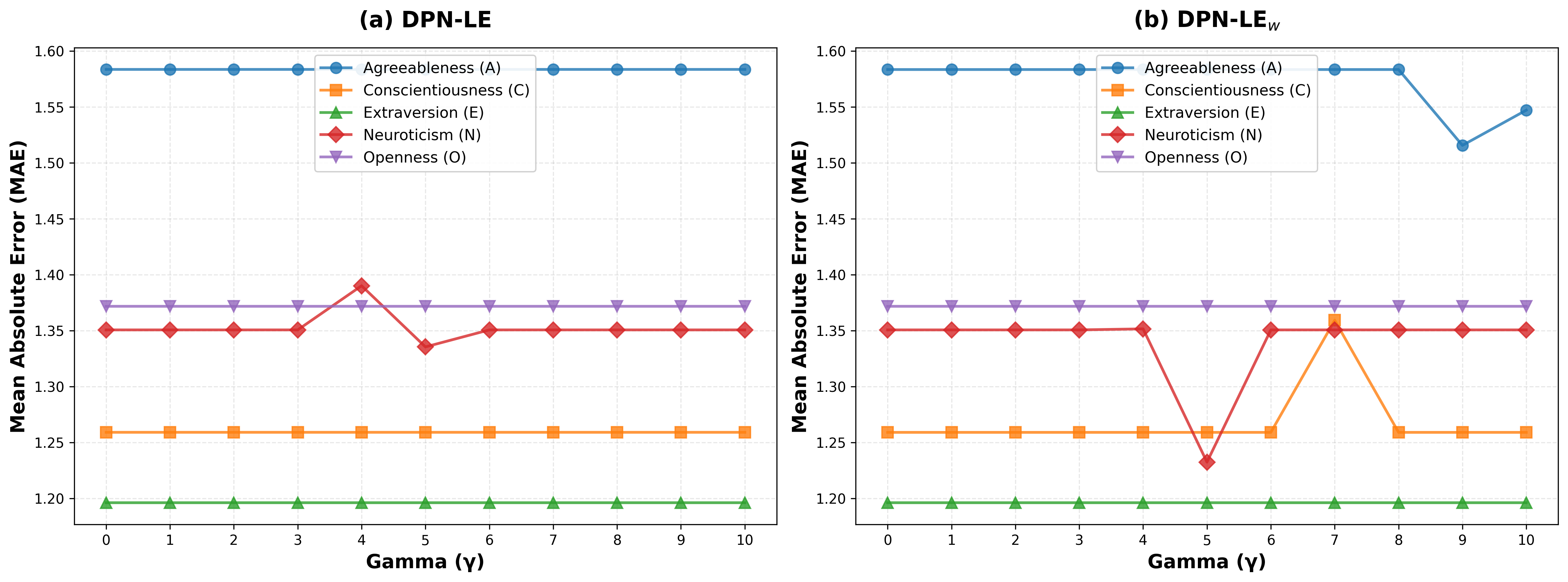}
\caption{MAE vs. intervention strength $\gamma$ on IPIP-NEO-300 test for both DPN-LE variants across all Big Five traits. Lower MAE indicates better personality alignment. (a) DPN-LE shows trait-specific optimal $\gamma$ ranges. (b) DPN-LE$_w$ exhibits smoother curves and more stable performance with reduced sensitivity to the intervention strength parameter due to the layer-wise weighting mechanism.}
\label{fig:mae-gamma-combined}
\end{figure*}

Tables~\ref{tab:ablation-gamma} and~\ref{tab:ablation-quantile} provide detailed numerical results for the ablation studies visualized in Figure~\ref{fig:ablation} of the main paper. Additionally, Figure~\ref{fig:mae-gamma-combined} shows the relationship between intervention strength $\gamma$ and Mean Absolute Error (MAE) on the IPIP-NEO-300 test for both DPN-LE variants across all Big Five traits.

\begin{table*}[ht]
\centering
\small
\begin{tabular}{c|cc|cc|cc|cc|cc|cc}
\toprule
\multirow{2}{*}{$\gamma$} & \multicolumn{2}{c|}{Agr.} & \multicolumn{2}{c|}{Con.} & \multicolumn{2}{c|}{Ext.} & \multicolumn{2}{c|}{Neu.} & \multicolumn{2}{c|}{Ope.} & \multicolumn{2}{c}{Avg.} \\
\cmidrule{2-13}
& T & F & T & F & T & F & T & F & T & F & T & F \\
\midrule
\multicolumn{13}{c}{\textit{DPN-LE}} \\
\midrule
0.5 & 6.84 & 9.96 & 6.97 & 10.00 & 7.33 & 9.99 & 8.62 & 9.98 & 6.86 & 9.98 & 7.32 & 9.98 \\
0.8 & 7.76 & 9.89 & 7.43 & 9.94 & 7.99 & 9.89 & 9.32 & 9.94 & 7.62 & 9.59 & 8.02 & 9.85 \\
1.0 & 8.28 & 9.76 & 8.15 & 9.73 & \textbf{8.39} & 9.20 & 9.62 & 9.75 & 8.49 & 8.22 & 8.59 & 9.33 \\
1.2 & 8.77 & 9.56 & 8.61 & 9.00 & 8.07 & 7.61 & \textbf{9.85} & 9.10 & \textbf{8.97} & 5.56 & 8.85 & 8.17 \\
1.5 & \textbf{9.51} & 8.74 & \textbf{9.19} & 6.84 & 6.20 & 2.67 & 9.71 & 5.35 & 7.82 & 3.52 & 8.49 & 5.42 \\
\midrule
\multicolumn{13}{c}{\textit{DPN-LE$_w$}} \\
\midrule
0.5 & 6.63 & 9.99 & 6.96 & 10.00 & 7.03 & 10.00 & 8.26 & 10.00 & 6.70 & 9.99 & 7.12 & 10.00 \\
0.8 & 7.27 & 9.90 & 7.53 & 9.98 & 7.82 & 9.97 & 9.16 & 9.95 & 7.42 & 9.67 & 7.84 & 9.89 \\
1.0 & 8.07 & 9.81 & 7.84 & 9.83 & 8.13 & 9.64 & 9.52 & 9.89 & 8.11 & 8.96 & 8.33 & 9.63 \\
1.2 & 8.41 & 9.69 & 8.35 & 9.19 & \textbf{8.25} & 8.94 & 9.69 & 9.44 & \textbf{8.72} & 6.94 & 8.68 & 8.84 \\
1.5 & \textbf{9.21} & 9.13 & \textbf{9.15} & 7.84 & 6.41 & 3.91 & \textbf{9.91} & 7.68 & 8.62 & 4.34 & 8.66 & 6.58 \\
\bottomrule
\end{tabular}%
\caption{Ablation study on intervention strength $\gamma$ (fixed Q995, Cohen's d $\geq$ 0.8). T = Trait Total (↑), F = Fluency Total (↑). Best T scores are \textbf{bold}.}
\label{tab:ablation-gamma}
\end{table*}

\begin{table*}[ht]
\centering
\small
\begin{tabular}{c|cc|cc|cc|cc|cc|cc}
\toprule
\multirow{2}{*}{Quantile (\%)} & \multicolumn{2}{c|}{Agr.} & \multicolumn{2}{c|}{Con.} & \multicolumn{2}{c|}{Ext.} & \multicolumn{2}{c|}{Neu.} & \multicolumn{2}{c|}{Ope.} & \multicolumn{2}{c}{Avg.} \\
\cmidrule{2-13}
& T & F & T & F & T & F & T & F & T & F & T & F \\
\midrule
\multicolumn{13}{c}{\textit{DPN-LE}} \\
\midrule
Q999 (0.1\%) & 6.82 & 9.97 & 7.06 & 9.99 & 7.80 & 9.91 & 8.81 & 9.91 & 7.27 & 9.71 & 7.55 & 9.90 \\
Q995 (0.5\%) & 8.28 & 9.76 & 8.15 & 9.73 & \textbf{8.39} & 9.20 & 9.62 & 9.75 & 8.49 & 8.22 & 8.59 & 9.33 \\
Q990 (1.0\%) & 8.62 & 9.74 & 8.45 & 8.96 & 7.09 & 7.03 & \textbf{9.72} & 9.56 & \textbf{9.12} & 7.03 & 8.60 & 8.46 \\
Q980 (2.0\%) & 8.90 & 9.51 & 8.46 & 9.21 & 7.20 & 5.64 & \textbf{9.72} & 9.17 & 8.87 & 7.14 & 8.63 & 8.13 \\
Q970 (3.0\%) & \textbf{9.00} & 9.42 & \textbf{8.62} & 9.11 & 7.22 & 5.16 & 9.70 & 8.45 & 8.87 & 6.77 & 8.68 & 7.78 \\
\midrule
\multicolumn{13}{c}{\textit{DPN-LE$_w$}} \\
\midrule
Q999 (0.1\%) & 6.81 & 9.97 & 7.06 & 10.00 & 7.49 & 9.99 & 8.69 & 9.97 & 7.01 & 9.81 & 7.41 & 9.95 \\
Q995 (0.5\%) & 8.07 & 9.81 & 7.84 & 9.83 & \textbf{8.13} & 9.64 & 9.52 & 9.89 & 8.11 & 8.96 & 8.33 & 9.63 \\
Q990 (1.0\%) & 8.40 & 9.76 & 8.26 & 9.30 & 7.72 & 8.34 & 9.52 & 9.66 & 8.51 & 8.44 & 8.48 & 9.10 \\
Q980 (2.0\%) & 8.50 & 9.62 & 8.11 & 9.49 & 7.52 & 7.15 & 9.53 & 9.59 & \textbf{8.63} & 8.33 & 8.46 & 8.84 \\
Q970 (3.0\%) & \textbf{8.73} & 9.59 & \textbf{8.28} & 9.44 & 7.13 & 5.60 & \textbf{9.56} & 9.35 & 8.56 & 8.02 & 8.45 & 8.40 \\
\bottomrule
\end{tabular}
\caption{Ablation study on quantile threshold (fixed $\gamma$=1.0, Cohen's d $\geq$ 0.8). Percentages indicate the proportion of neurons selected per layer. T = Trait score ($\uparrow$), F = Fluency ($\uparrow$). Best T scores are \textbf{bold}.}
\label{tab:ablation-quantile}
\end{table*}

\subsection{Neuron Selection Statistics}
Table~\ref{tab:neuron-stats} shows the number of neurons selected by DPN-LE per layer for both models under Q995, using the model-specific Cohen's $d$ thresholds in Table~\ref{tab:hyperparams}. Both configurations select approximately 0.5\% of total MLP neurons per layer.

\section{Prompt Templates}
\label{sec:appendix-prompts}

\subsection{Big Five Trait Descriptions}

Table~\ref{tab:trait-descriptions} provides representative examples of personality descriptions used for generating contrastive samples. The PersonalityBench dataset contains 80 high-trait and 80 low-trait descriptions per trait; we show condensed summaries that capture the key characteristics of each direction.

\begin{table*}[ht]
\centering
\begin{tabular}{p{2.5cm}p{6.5cm}p{6.5cm}}
\toprule
\small
\textbf{Trait} & \textbf{High Expression} & \textbf{Low Expression} \\
\midrule
Openness & Creative, curious, appreciates art and new experiences, imaginative, open to unconventional ideas & Practical, conventional, prefers routine and familiarity, down-to-earth, traditional \\
\addlinespace
Conscientiousness & Organized, disciplined, goal-oriented, reliable, thorough, plans ahead carefully & Spontaneous, flexible, casual about obligations, adaptable, prefers improvisation \\
\addlinespace
Extraversion & Outgoing, energetic, talkative, enjoys social interactions, seeks excitement and stimulation & Reserved, quiet, prefers solitude, reflective, comfortable with smaller social circles \\
\addlinespace
Agreeableness & Cooperative, trusting, helpful, empathetic, considerate of others' feelings & Competitive, skeptical, challenging, direct, prioritizes own interests \\
\addlinespace
Neuroticism & Emotionally reactive, prone to stress and anxiety, experiences mood swings & Emotionally stable, calm under pressure, resilient, even-tempered \\
\bottomrule
\end{tabular}
\caption{Representative Big Five personality trait descriptions for high and low expressions.}
\label{tab:trait-descriptions}
\end{table*}

\end{document}